%% file: main.tex
\documentclass[conference]{IEEEtran}
\usepackage{times}

\usepackage[numbers]{natbib}
\usepackage{multicol}
\usepackage[bookmarks=true]{hyperref}

\usepackage{graphicx, caption, subcaption, amsmath}
\usepackage{array, multirow}
\usepackage{xcolor}
\graphicspath{ {figures/} }

\usepackage{array}
\newcolumntype{P}[1]{>{\centering\arraybackslash}p{#1}}
\newcolumntype{M}[1]{>{\centering\arraybackslash}m{#1}}

\newcommand{\vect}[1]{\boldsymbol{#1}}

\DeclareMathOperator*{\argmin}{arg\,min}

\begin{document}

\title{High Fidelity Capture, Reconstruction, and Transfer of Human Demonstrations for Robot-Assisted Bathing}

\IEEEoverridecommandlockouts 

\author{
    Arjun S. Lakshmipathy$^{1}$, 
    Jonathan P. King$^{1}$, 
    Ethan Zuo$^{1*}$, 
    Rohit Satishkumar$^{1*}$, \\ 
    Hongyi Chen$^{1}$, 
    Jeffrey Ichnowski$^{1}$, 
    Dan Ding$^{2}$, 
    Zackory Erickson$^{1}$, 
    and Nancy S. Pollard$^{1}$
    \thanks{$^{1}$School of Computer Science, Carnegie Mellon University}%
    \thanks{$^{2}$School of Health and Rehabilitation Sciences, University of Pittsburgh}%
    \thanks{$^{*}$Denotes equal contribution}%
}

\maketitle

\input{sections/abstract}

\IEEEpeerreviewmaketitle

\input{sections/introduction}

\input{sections/related_work}

\input{sections/methods}

\input{sections/experiments_and_results}

\input{sections/discussion}

\input{sections/conclusion}

\input{sections/acknowledgements}

\bibliographystyle{unsrt}
\bibliography{references}

\clearpage

\input{sections/appendix}

\end{document}

%% file: sections/abstract.tex
\begin{abstract}

Despite the demand for robots in high-value clinical tasks like bathing, contemporary systems still lack the safety and reliability required for complex, sustained physical interaction with humans. A key challenge hindering the development of such systems is that collecting, understanding, and effectively transferring highly dynamic, contact-rich human bathing demonstrations is difficult, even with modern motion and tactile sensing equipment. We present a straightforward, but effective framework for doing so with high fidelity by utilizing contact regions as a key processing primitive. We use our framework to build a dataset of bathing demonstrations performed by trained clinicians on human subjects. We then use this dataset to design and control an arm-mounted dexterous soft hand to perform bathing tasks on a mannequin using open- and closed-loop strategies. Our dataset is the first to provide high quality synchronized motion, shape, contact, and force during sustained, contact-rich human-human interaction, and our transfer strategies demonstrate effective use of these data across multiple levels of the robotics stack. All relevant materials will be publicly released to enable further advancements in physical human-robot interaction (pHRI) research.

\end{abstract}

%% file: sections/introduction.tex
\section{Introduction}

Bathing is an essential activity of daily living and important component of long-term personal hygiene~\cite{dunlop1997disability}. The Katz scale~\cite{katz1963studies} identifies the ability to bathe as one of six major categories that quantify an individual's level of independence; consequently, losing this ability significantly impacts quality of life. As the number of Americans with bathing-related impairments rises and the pool of trained caregivers shrinks~\cite{gill2006epidemiology}, the demand for bathing assistance is rapidly outstripping the available supply of care.

Robot-assisted bathing solutions have the potential to significantly expand access to care while reducing existing caregiver burden~\cite{czuba2012ergonomic,darragh2015musculoskeletal}. However, building systems that can safely and reliably perform the sustained, highly dynamic, and contact-rich human interactions inherent to bathing is technically challenging. While several impressive systems have emerged recently~\cite{erickson2019multidimensional,liu2022characterization,madan2024rabbit}, they generally rely on idealized definitions of bathing ranging from best practice guidelines~\cite{madan2024rabbit} to coarse-grained performance metrics (e.g. debris removal percentages~\cite{huang2022soft,liu2022characterization}) rather than real data from human demonstrations. Consequently, many important details such as realistic hand movements and shape adjustments across different body segments, reactive corrections in response to patient movement, the evolution of contact regions between the hand and body over time, and forces applied over those regions are not considered. To date, there are no comprehensive data that capture how caregivers actually bathe people.

\begin{figure}[h!]
\centering
\includegraphics[trim={0 0 0 0},clip,width=1.0\linewidth]{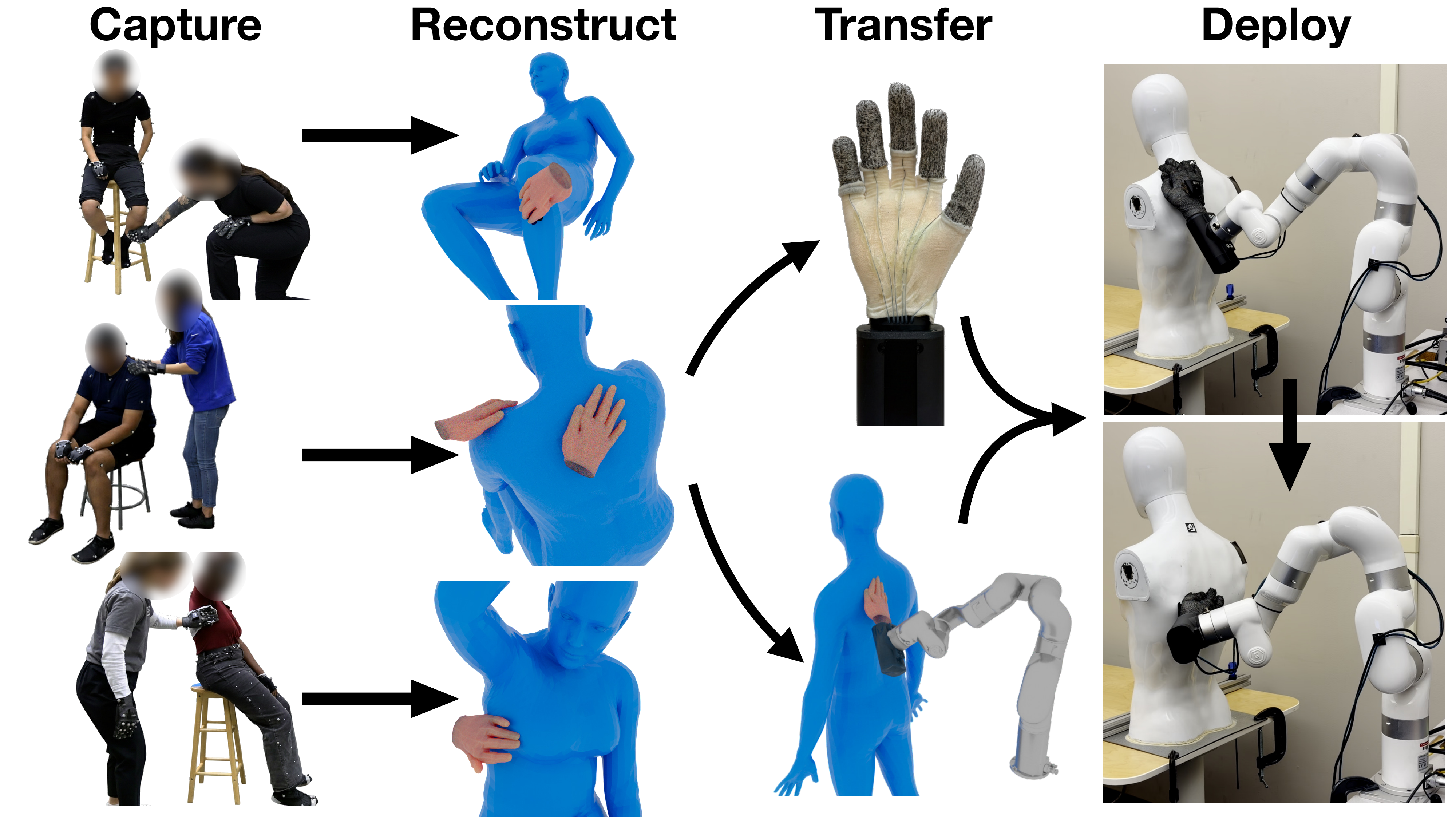}
\caption{Our contributions enable straightforward, yet high quality reconstructions of human bathing demonstrations along with strategies for robot skill transfer.}
\label{fig:papersummary}
\end{figure}

This paper presents a straightforward, but effective approach for high fidelity capture and reconstruction of human bathing demonstrations, as well as techniques that utilize this information to facilitate the design and control of robot systems. Our capture pipeline takes full advantage of modern motion and tactile capture systems, but crucially provides solutions and guidance to address issues that come with reconstructing fine-grained physical interaction. Central to our techniques is the use of contact regions as a key processing primitive, which we demonstrate naturally addresses many pain points in the pipeline. We make the following contributions:

\begin{itemize}
    \item A high quality reconstruction pipeline for time-synchronized motion, shape, contact, and force of real human-assisted bathing demonstrations
    \item A dataset of 128 captures and 257,000 total frames sourced across three clinicians and six human subjects
    \item Design and control strategies for an anthropomorphic tendon-driven soft hand using our dataset
    \item A retargeting pipeline to transfer demonstrations to a composite arm-hand system 
\end{itemize}

Our contributions enable visualization, understanding, and translation of human-assisted bathing demonstrations at an unprecedented level of detail, which we anticipate will be of considerable interest to the clinical and pHRI communities.

%% file: sections/related_work.tex
\section{Related Work}

We discuss related work spanning robot-assisted bathing systems, tactile-augmented human motion capture and transfer, and contract-driven strategies for grasping and manipulation.

\subsection{Robot-Assisted Bathing Systems}

Multiple robot-assisted bathing platforms have previously been proposed, ranging from fixed installations to mobile manipulators. Fixed installations~\cite{dometios2017real,zlatintsi2018multimodal} are often custom-engineered for specific robotic platforms. As a result, bathing tasks and motion planning must be developed from scratch for each system and location, precluding the direct transfer of skills from human demonstrations. Alternative approaches utilize mobile manipulators~\cite{king2010towards,madan2024rabbit,liu2022characterization,erickson2019multidimensional} to bypass location constraints, though the specific methodologies employed vary. Liu et. al.~\cite{liu2022characterization} propose a unique wearable design optimized for debris cleanup. King et. al.~\cite{king2010towards}, Erickson et. al.~\cite{erickson2019multidimensional}, and most recently Madan et. al.~\cite{madan2024rabbit} use more traditional designs and instead focus their efforts on perception challenges inherent in bathing tasks. However, these works prioritize the development of robot-specific policies over the capture and transfer of human demonstrations. We focus on the latter, investigating how human skills can be effectively captured, reconstructed, and translated to robotic platforms.

\subsection{Tactile-Augmented Human Motion Capture and Transfer}

Recent research has increasingly focused on jointly capturing motion and force information from human demonstrations, as well as transferring skills to robots. A wide array of sensing modalities have been investigated to date, ranging from force-sensitive resistive~\cite{sundaram2019learning,luo2024adaptive} and capacitive arrays~\cite{ppsTactileGlove} to more specialized approaches such as embedded-camera-based~\cite{fang2025dexop}, magnetic~\cite{adeniji2025feel,yin2025osmo}, acoustic~\cite{mao2025visuo}, and Fiber Bragg Grating sensors~\cite{xing2025taccap}. Skill transfer strategies, which are predominantly learning-based, include using demonstrations to train transformers~\cite{adeniji2025feel} or condition diffusion policies~\cite{yin2025osmo}. Our work extends this literature to pHRI, which is a challenging domain characterized by sustained, predominantly sliding contact-rich interactions across the imprecise and shifting geometry of the human body. Concretely, we utilize capacitive sensing for our captures but focus more on data processing and straightforward methods of transfer at multiple layers of the robotics stack. 

\subsection{Contact-Driven Strategies for Grasping and Manipulation}

Contact information is fundamental to grasping and manipulation; it is frequently leveraged as a loss term for pose optimization~\cite{lakshmipathy2023contactedit,turpin2022grasp,brahmbhatt2019contactgrasp}, a constraint for training generative models~\cite{christen2022d,wu2022saga}, a prior for retargeting human trajectories~\cite{lakshmipathy2025kinematic,mandi2025dexmachina,pan2025spider}, and a primitive for motion planning~\cite{pang2023global,cheng2022contact}. Modern high quality human motion and interaction datasets~\cite{ taheri2020grab,fan2023arctic,delpreto2022actionsense,song2025opentouch}, many of which serve as the backbones and testbeds of manipulation research efforts, are notably prioritizing the inclusion of detailed contact information. We take these efforts one step further by showing that high fidelity bathing reconstruction and transfer are \textit{critically dependent} on the effective use of contact areas as a processing primitive.

%% file: sections/methods.tex
\section{Methods}

\begin{figure*}
\centering
\includegraphics[trim={0 0 0 0},clip,width=1.0\linewidth]{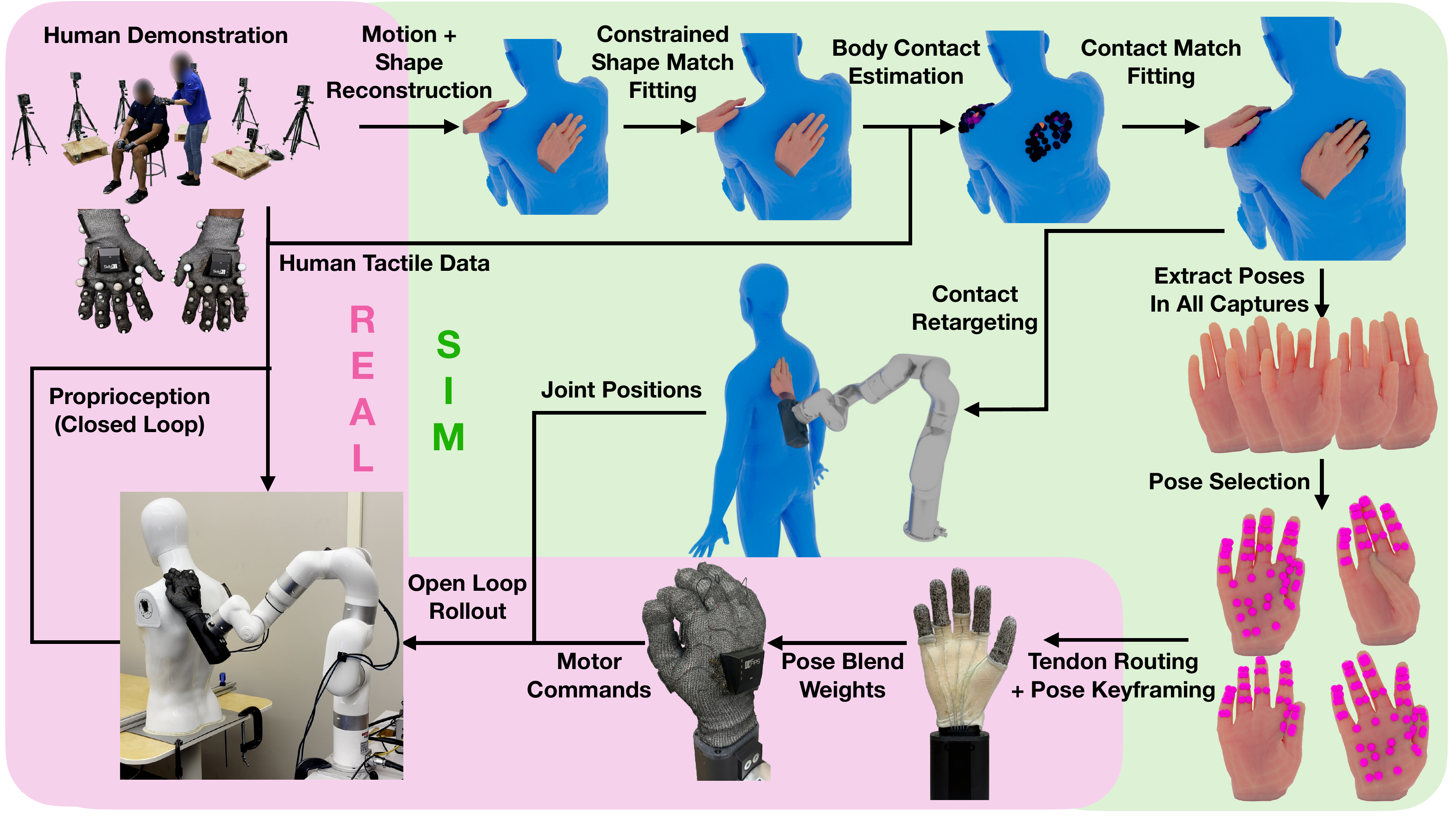}
\caption{Overview of our pipeline. Starting with captures of motion and tactile forces from human bathing demonstrations, we digitally reconstruct time-synchronized data across shape, movement, and contact force with high accuracy. We then utilize these reconstructions to both design and control an anthropomorphic soft hand, as well as retarget results to a robot arm. We demonstrate the final system's efficacy by replicating bathing tasks on a mannequin in the real world. A core contribution of this work is the systematic integration of contact distributions and force data throughout the entire pipeline. Pink and green partitions delineate steps performed in the real world and simulation respectively.}
\label{fig:overview}
\end{figure*}

Figure~\ref{fig:overview} provides an overview of our capture, reconstruction, and transfer pipeline. We detail each component below.

\subsection{Data Capture}

\begin{figure}
\centering
\includegraphics[trim={0 0 0 0},clip,width=1.0\linewidth]{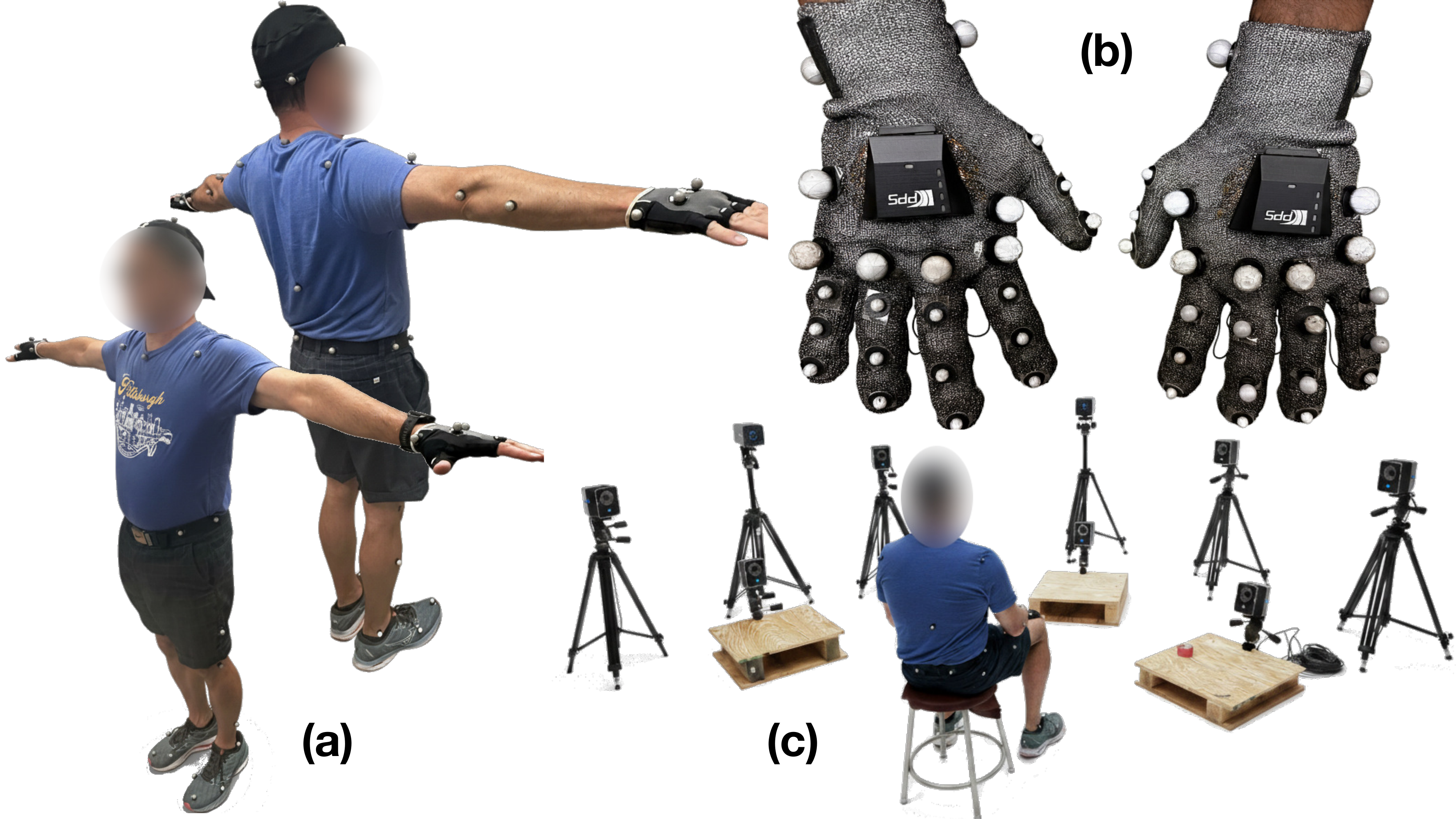}
\caption{Optical marker configurations used on (a) human subjects and (b) tactile force gloves. (c) Cameras are floor mounted and positioned in a semi-circle configuration.}
\label{fig:mocapequipment}
\end{figure}

\begin{figure}
\centering
\includegraphics[trim={0 0 0 0},clip,width=1.0\linewidth]{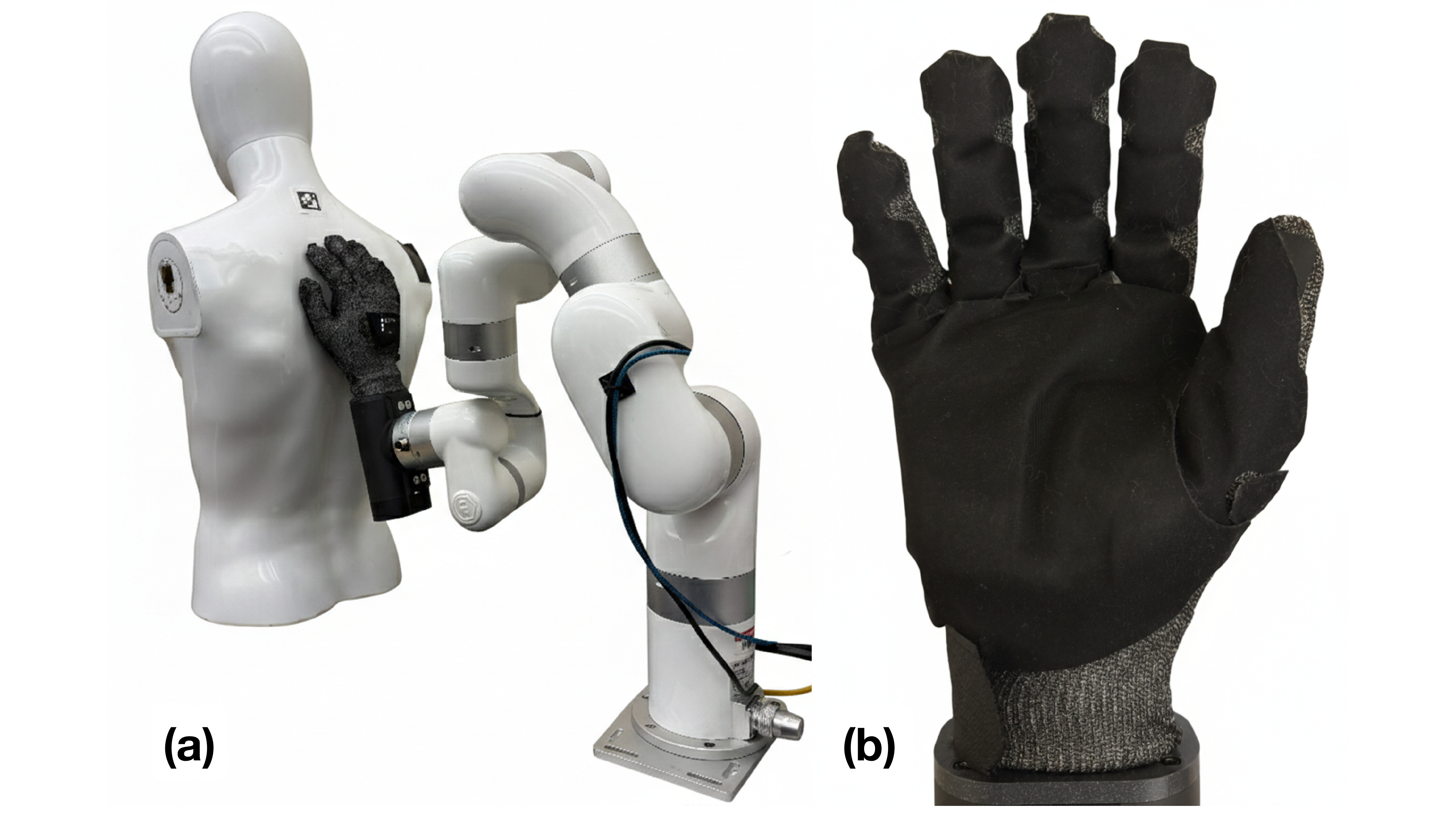}
\caption{(a) The robot-assisted bathing setup, consisting of a robotic arm~\cite{ufactoryXarm7}, DexKit robotic hand~\cite{king2025dexkit}, and mannequin. (b) The DexKit hand is outfitted with a tactile sensing glove~\cite{ppsTactileGlove} with grip material~\cite{3MGripTape} coating the palmar side to prevent slip and moisture damage.}
\label{fig:deploymentequipment}
\end{figure}

We collected demonstrations from three trained clinicians and six subjects. Each clinician was randomly assigned to two subjects. Subjects were required to be of at least 18 years of age and able-bodied, while clinicians were required to have prior experience with assisted bathing. Subjects were outfitted with 52 optical markers for motion tracking and encouraged, but not required, to wear dark summer clothing to minimize optical reflectance while exposing as many extremities as possible for direct skin contact. Clinicians were instructed to wear a pair of commercially available tactile sensing gloves~\cite{ppsTactileGlove} and perform bathing motions on the subject's skin, or alternatively clothing if skin exposure was not possible, using a hypoallergenic wet wipe. The dorsal side of each glove was outfitted with 23 optical markers, while the palmar side was outfitted with a layer of adhesive-backed water-resistant grip material~\cite{3MGripTape} to protect the sensor electronics from incidental moisture damage from the wipe. The grip material was laser cut into a ``tabbed" pattern designed to maximize surface coverage with minimal impact to range-of-motion (ROM). Notably, we found that the grip material provided the best compromise between water resistance, adhesion, and deformability in comparison to other material candidates (e.g. vinyl, silicone). Figures~\ref{fig:mocapequipment} and~\ref{fig:deploymentequipment} illustrate our equipment.

The sensory array of each glove consists of 65 distinct ``taxels", or pressure-sensitive pads, distributed throughout the palmar and digital surfaces. The sensors measure changes in pressure via differential capacitance~\cite{baxter1997capacitive} relative to a baseline taken at the start of each collection. We assume all forces are normal to each taxel.

Subjects were positioned upright on a stool surrounded by a set of 20 Vicon Vantage-V16 cameras~\cite{viconV16Specs}. Clinicians were asked to bathe subjects in the same way as they would a patient in a medical or assisted-care facility. Markers inhibiting bathing motions of specific body parts were temporarily removed during captures of those parts.

Each capture consisted of the tracked 3D positions of all subject markers and glove optical markers, as well as pressure readings from all 65 taxels per glove. Tactile and marker data were streamed at 10 and 120 Hz respectively and synchronized using Vicon timecodes. All subjects and clinicians were also asked to perform ROM tests prior to their first capture. We ensure all markers are labeled and have unbroken trajectories.

\subsection{Shape and Motion Reconstruction}

We convert tracked marker positions and force measurements into complete reconstructions of shape, motion, and contact using a multi-stage pipeline. We first utilize the SMPL-X~\cite{pavlakos2019expressive} and MANO~\cite{romero2017mano} parametric models to estimate each subject's body and each clinician's hand shape parameters from their respective ROM tests using a random subset of 50 frames, as well as canonical marker position estimates relative to the baseline shape. We then estimate all pose and corrective blend shape parameters for each bathing capture trajectory using the MoSh++ solver~\cite{mahmood2019moshpp}. This process effectively converts point-based marker trajectories into complete mesh trajectories of the body and hands.

\begin{figure}
\centering
\includegraphics[trim={0 5cm 0 9cm},clip,width=1.0\linewidth]{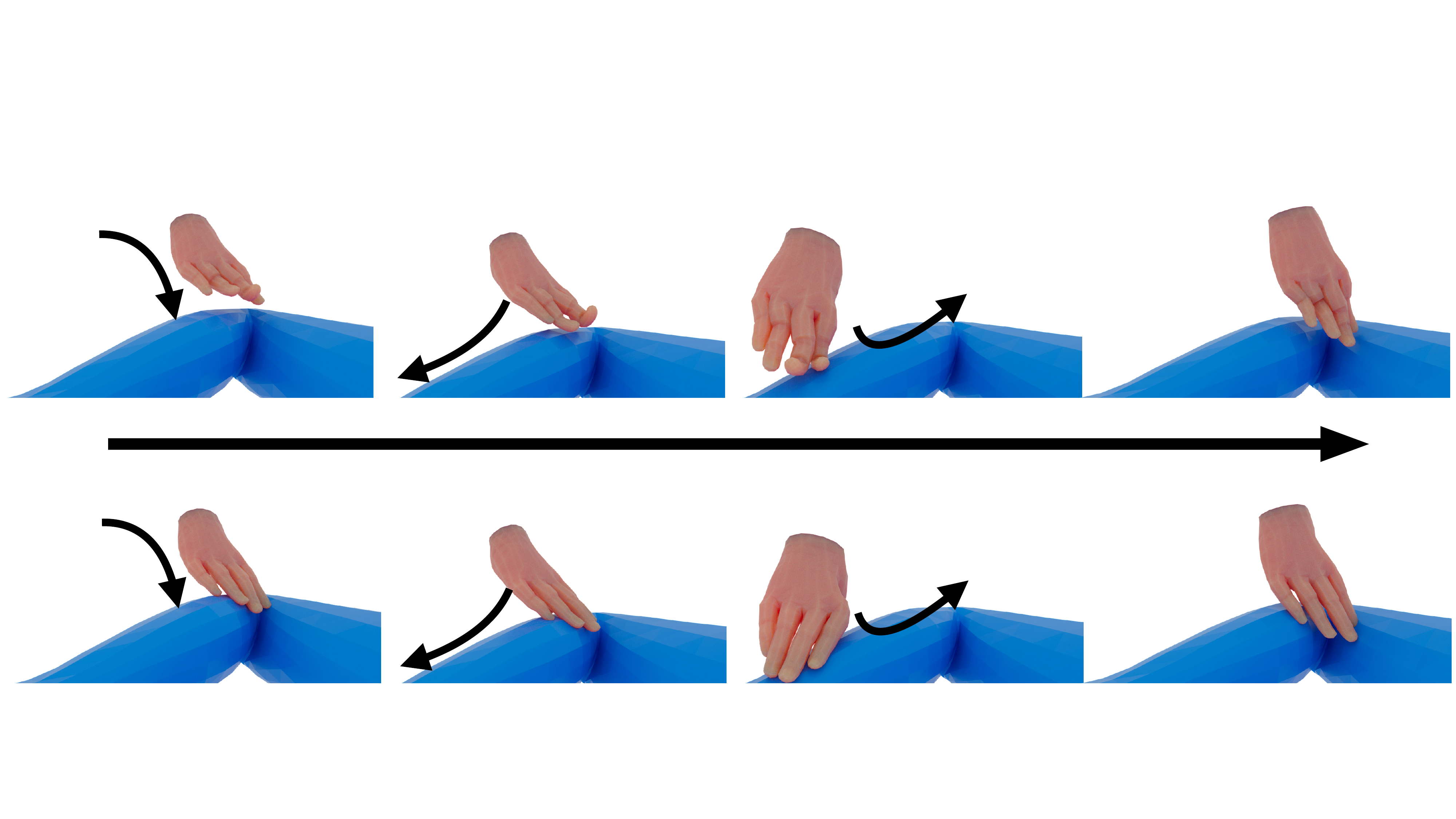}
\caption{Comparison film strip of a leg bathing reconstruction using (top) MoSh++~\cite{mahmood2019moshpp} and (bottom) our reconstruction pipeline. Black arrows illustrate subsequent movement directions. Our method effectively resolves finger contortion and gap artifacts without any significant parameter tuning.}
\label{fig:moshartifacts}
\end{figure}

However, as illustrated in Figure~\ref{fig:moshartifacts}, the process routinely produces serious artifacts such as highly contorted finger poses and major gaps of separation between the hands and body, rendering it inadequate for high quality bathing reconstruction. These artifacts largely result from the necessary process of fitting bare-skin hand shape parameters to gloved captures. To mitigate these inherent discrepancies, we implement a series of straightforward, yet robust solutions.

First, to combat finger contortion artifacts, we replace the baseline fitted hand mesh with a constrained skeleton-driven reduced-DOF model that honors anatomical joint limits~\cite{magee2013orthopedic}. We use the same baseline shape parameters and drive the constrained mesh using linear blend skinning~\cite{magnenat1989joint}. Joint angles are computed per frame using the following objective:

\begin{equation}
    \begin{aligned}
        \displaystyle \vect{\theta}^* = \argmin_{\vect{\theta}} \quad & \lambda_{s} \sum_{i=0}^{V} ||v_{bi} -v_{ci}||_2^2 + \lambda_{j} ||\vect{\theta} - \vect{\theta_{p}}||_2^2\\
        \textrm{s.t.} \quad & \vect{\theta_L} \leq \vect{\theta} \leq \vect{\theta_U}
    \end{aligned}
    \label{eq:optframe}
\end{equation}

\noindent where $v_{bi}$ and $v_{ci}$ represent corresponding vertices of the baseline and constrained hand meshes respectively, $V$ is the total number of MANO hand vertices, $\vect{\theta}$ is the constrained hand DOF vector, $\vect{\theta_L}$ and $\vect{\theta_U}$ are the lower and upper DOF bounds, $\vect{\theta_{p}}$ is the existing pose, and $\lambda_{s}$ and $\lambda_{j}$ are weighting hyperparameters. We adopt an existing motion synthesis pipeline to solve for the full trajectory $\vect{\Theta}^*$ of each constrained hand based on the per-frame pose estimates~\cite{lakshmipathy2025kinematic}, and refer to this fitting process as the \textit{shape match} pass.

We next utilize contact data to combat the gap artifacts. Starting from one-time calibrated hand taxel position estimates, we estimate corresponding body contact regions per frame using closest-point queries~\cite{sawhneyFCPW} and store the trajectories in barycentric coordinates. Next, utilizing tactile force data, we  binarize the signal into activations and filter out all inactive contacts. We then perform a second \textit{contact match} pass by adding a single additional objective term to Eq.~\ref{eq:optframe}:

\begin{equation}
    \lambda_{c} \sum_{i=0}^{C} \left\| c_{hi} - c_{bi} \right\|_2^2
\end{equation}

\noindent where $c_{hi}$ and $c_{bi}$ represent corresponding contact points on the constrained hand and body meshes respectively, $C$ represents the total number of taxels per hand, and $\lambda_{c}$ is a weighting hyperparameter. Final reconstructions are complete following the contact pass. We use hyperparameters $\lambda_{s}=1,\lambda_{j}=50$ and $\lambda_{c}=40$ for all captures and 100 iterations per optimization. We select these relative weights to offset dimensionality differences between $V$ (776), $C$ (65 / hand), and $||\vect{\theta}||$ (29).

\subsection{Design and Control of a Tendon-Driven Soft Hand}

We select the anthropomorphic DexKit platform~\cite{king2025dexkit} for demonstration transfer primarily due to its compliance and secondarily for its customizability. However, because the hand is fully soft with potentially infinite DOFs and unknown kinematics, three problems must be solved to use it effectively: (1) determining a tendon routing capable of achieving desired poses, (2) determining motor positions to actuate the tendons accordingly, and (3) building an effective control space. The combination of shape, motion, and contact offered by our reconstructions helps solve each of these problems.

Starting from a dataset of human hand poses $P$, we first select a strictly smaller subset of candidates $S \subset P$ that sufficiently covers the full space. We do so by solving an NP-hard maximum diversity problem~\cite{kuo1993analyzing} via greedy approximation~\cite{ghosh1996computational}. We then manually select a subset $S' \subset S$ that can be physically mimicked on DexKit under a given motor budget and commit to a tendon routing capable of generating the mimicked poses.

\begin{figure}
\centering
\includegraphics[trim={0 0 0 0},clip,width=1.0\linewidth]{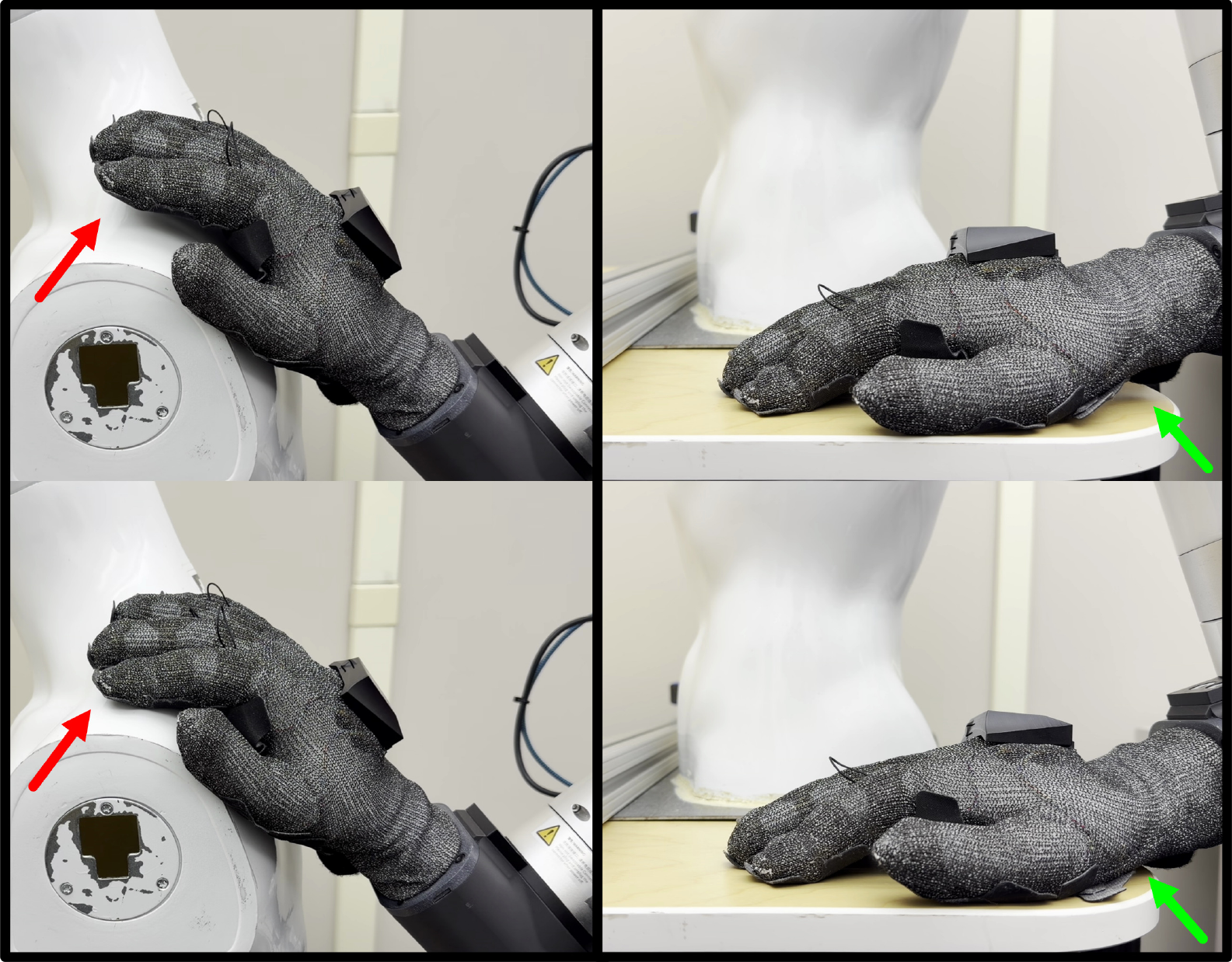}
\caption{Visualizations of improved surface contact following hand pose correction, comparing the original (\textit{top}) and corrected (\textit{bottom}) states. In the first example (\textit{left}), the red arrows highlight a significant increase in finger contact. In the second (\textit{right}), the green arrows indicate the hand flattening against the table as the heel moves closer to the surface.}
\label{fig:glovecontactjustification}
\end{figure}

Next, we determine motor commands $\vect{\phi}$. As shown in Figure~\ref{fig:glovecontactjustification}, calculating motor commands that adapt to contact-induced pose corrections depends on identifying hand contact regions and accurately simulating those interactions. We can directly obtain these contact regions from the dataset.

We construct two matrices: $\vect{M_{S'}}$, where each column corresponds to a pose $s \in S'$, and $\vect{M_{\phi(S')}}$, where each column consists of the manually determined motor positions $\vect{\phi(s)}$ required to generate that pose. For a given pose $p \in P$, we solve the following constrained least-squares QP to obtain normalized linear blend weights $\vect{\beta}^*$:

\begin{equation}
    \begin{aligned}
        \displaystyle \vect{\beta}^* = \argmin_{\vect{\beta}} \quad & (\vect{M_{S'}}\vect{\beta} - p)^T(\vect{M_{S'}}\vect{\beta} - p)\\
        \textrm{s.t.} \quad & 0 \leq \vect{\beta} \leq 1
    \end{aligned}
    \label{eq:qp}
\end{equation}

We then compute the corresponding motor positions from the optimal blend weights as $\vect{\phi} = \vect{M_{\phi(S')}}\vect{\beta}^*$. Additional details of pose selection metrics, as well as final motor configuration and tendon routings, are available in the appendix.

\subsection{Arm Motion Retargeting and Control}

We mount the DexKit hand onto an xArm 7~\cite{ufactoryXarm7} to perform bathing tasks on a mannequin. We deploy pre-computed motor trajectories $\vect{\phi}$ directly to the hand, but must retarget the human hand wrist trajectories to obtain arm joint angles $\vect{\psi}$. Unfortunately, we found that the commonly used strategy of using human wrist positions and orientations as end-effector targets~\cite{sivakumar2022robotic} did not work well due to the arm's restricted workspace and the fact that hand trajectories were highly dependent on each subject's body shape and reactive motion.

Instead, we compute $\vect{\psi}$ using retargeted body contacts. Specifically, we use the TRAM video pose estimator~\cite{wang2024tram} to fit SMPL-X parameters to the mannequin, and then directly roll out contact trajectories computed on the source subject. Because contact trajectories are stored in barycentric coordinates, the retargeting operation is agnostic to the target body's shape or pose parameters. We can thus always perform the rollout regardless of any shape or pose divergences between the source and target domain. Next, we replace the DexKit hand with the source MANO hand, remove all wrist DOFs, copy $\vect{\Theta}^*$ from the original motion, and freeze the hand into those poses. This step allows us to re-establish correspondences between the retargeted body contacts and original hand taxels, and subsequently solve for $\vect{\psi}$ using Eq.~\ref{eq:optframe}. The resulting trajectory $[\{\vect{\phi_0}, \vect{\psi_0}\}, \{\vect{\phi_1}, \vect{\psi_1}\}, ... , \{\vect{\phi_T}, \vect{\psi_T}\}]$ is then deployed open-loop on the real system using a modified version of the GELLO framework~\cite{wu2024gello}.

However, the open-loop controller frequently breaks, or alternatively applies too much, contact with the mannequin due to the gap between the real geometry and its fitted SMPL-X parameters in simulation. We create a closed-loop system by sensorizing the DexKit hand with the same tactile glove used in data capture, and utilize contact and force information from the capture to create a lower-level reactive controller. Specifically, at each time step, we apply a normalized corrective direction $N_C$ for the end-effector along the axis intersecting the end-effector and mannequin in the real world. We convert $\vect{\psi}$ to EE positions $FK(\vect{\psi})$ rolled out at a baseline frequency and apply corrective position updates at a fixed timestep $\Delta t$, creating inner loop EE position $FK(\vect{\psi}) + N_C\Delta t$. Corrective movements are applied toward the body until the online sum of tactile pressures matches the original measurement, then reversed if that value is exceeded.

%% file: sections/experiments_and_results.tex
\section{Experiments and Results}

We discuss experiments and results concerning our three major thrusts in the proceeding subsections.

\subsection{Dataset Statistics and Analysis}

\begin{figure}
\centering
\includegraphics[trim={0 7cm 0 5cm},clip,width=1.0\linewidth]{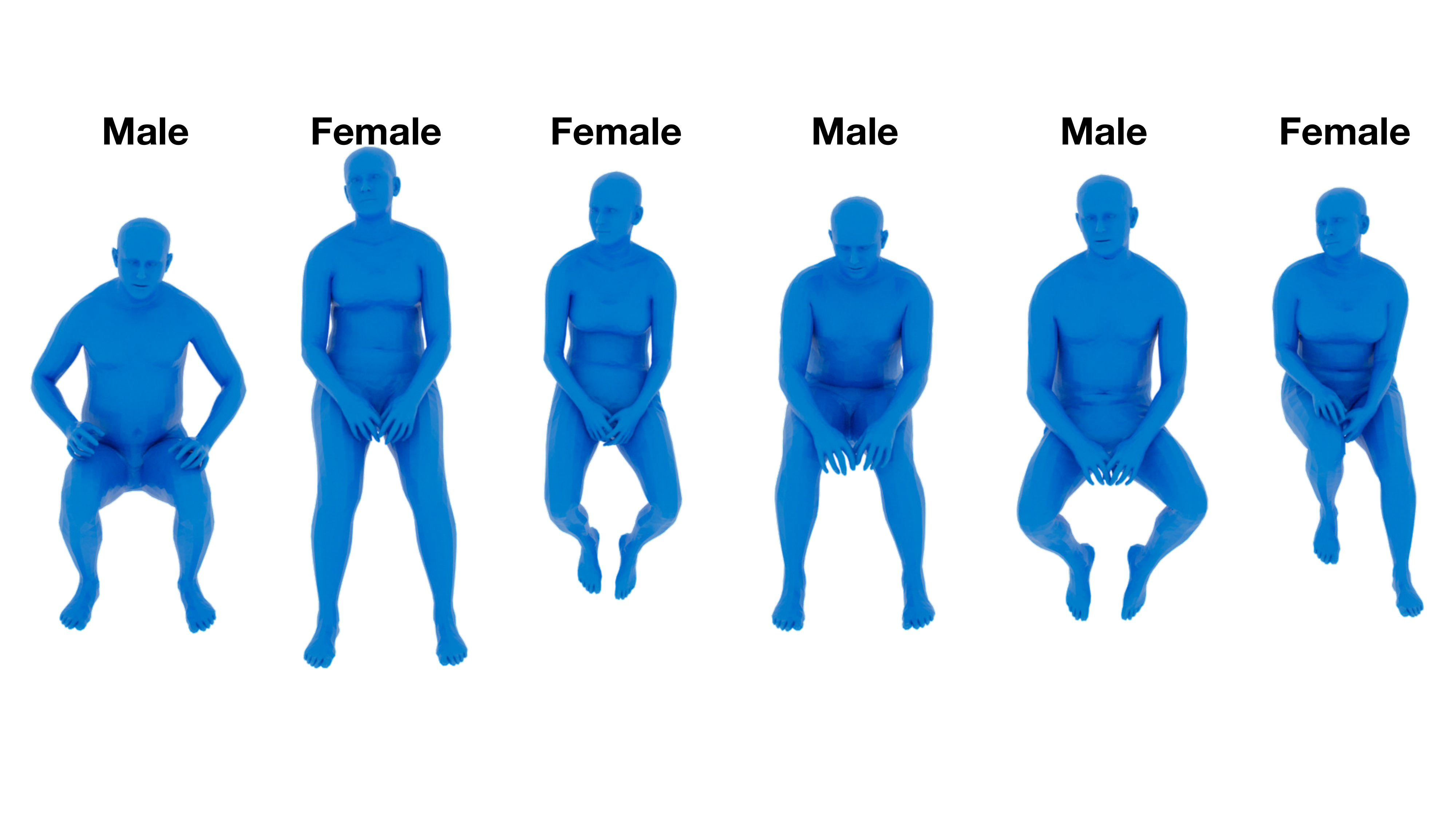}
\caption{MoSh++~\cite{mahmood2019moshpp} shape reconstructions of all subjects.}
\label{fig:subjectrenders}
\end{figure}

\begin{table}[h!]
\centering
\caption{Distribution of all captures in our dataset segmented by body part, assistance level, and pressure intensity.}
\begin{tabular}{ |>{\centering\arraybackslash}M{1.5cm}|M{1cm}|M{1.25cm}|M{1cm}|M{1.25cm}| }
    \hline
     & Limited, Mild & Limited, Strong & Full, Mild & Full, Strong \\
    \hline
    Right Arm & 10 & 8 & 9 & 7 \\
    \hline
    Right Leg & 6 & 6 & 3 & 3 \\
    \hline
    Right Thigh & 3 & 3 & N/A & N/A \\
    \hline
    Right Oblique & 2 & 2 & N/A & N/A \\
    \hline
    Left Arm & 7 & 6 & 7 & 6 \\
    \hline
    Left Leg & 6 & 6 & 4 & 3 \\
    \hline
    Left Thigh & 3 & 3 & N/A & N/A \\
    \hline
    Left Oblique & 2 & 2 & N/A & N/A \\
    \hline
    Back & 6 & 6 & N/A & N/A \\
    \hline
    Neck & 7 & 1 & N/A & N/A \\
    \hline
    Face & 1 & N/A & N/A & N/A \\
    \hline
\end{tabular}
\label{table:datasetdistribution}
\end{table}

Table~\ref{table:datasetdistribution} tabulates the distribution of captures across all subjects and clinicians in our dataset. Subjects ranged in age from 22 to 31 years, with an equal distribution of males and females. Renders of all reconstructed subject body shape parameters are available in Figure~\ref{fig:subjectrenders}. Clinicians were all female, aged 30 to 32 years. The average capture duration was 2007.34 frames ($\sigma = 536.615$).

Each capture is categorized by one of two assistance variables (limited or full) and one of two pressure intensities (mild or strong). Limited assistance captures assume subjects can independently lift and hold their limbs, while full assistance captures require clinicians to manually raise and lower them throughout the bathing process. Clinicians use their discretion to differentiate between mild and strong pressure, as no explicit guidelines are provided. We also note that not all variable permutations are available per body part. For example, face bathes only consist of mild pressure demonstrations due to the region's sensitivity, areas such as the beck, neck, and obliques cannot possibly require full support, and multiple clinicians opted to not provide full assistance for legs because the entire limb could already be accessed from a seated position. We acknowledge that bed-bathing may provide some additional demonstrations not represented in our dataset. Finally, while we cover most parts of the body, we do not have captures of hand bathing due to tracking difficulties or of the chest and pelvic regions due to IRB restrictions. Renders of representative results are available in the supplementary video. We provide additional metrics in the proceeding subsections.

\subsection{Visualizations of Contact Coverage}

\begin{figure}
\centering
\includegraphics[trim={0 5cm 0 5cm},clip,width=1.0\linewidth]{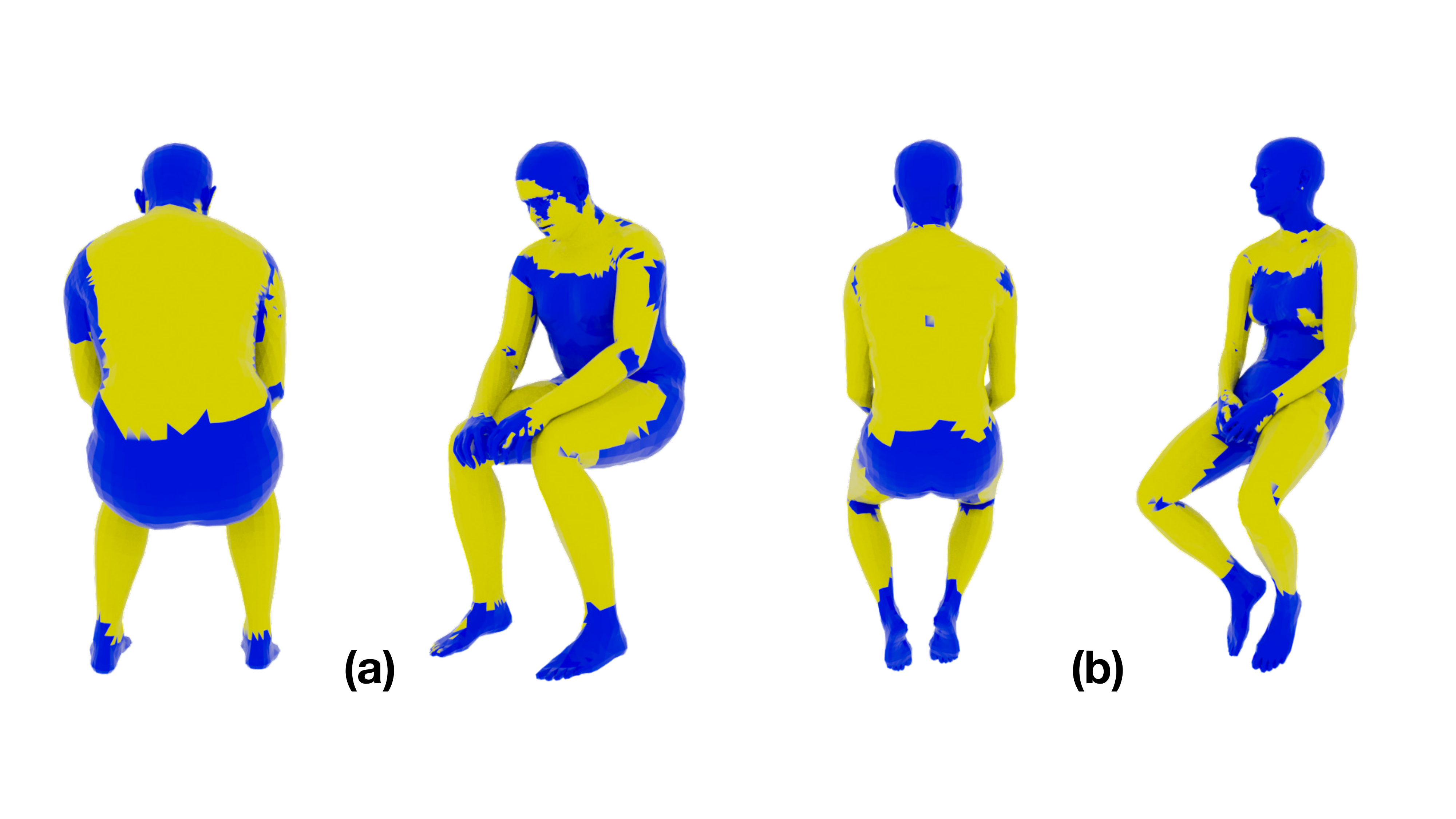}
\caption{Contact coverage (yellow) over all captures of a representative (a) male and (b) female subject.}
\label{fig:bodycoverage}
\end{figure}

\begin{figure}
\centering
\includegraphics[trim={0 5cm 0 5cm},clip,width=1.0\linewidth]{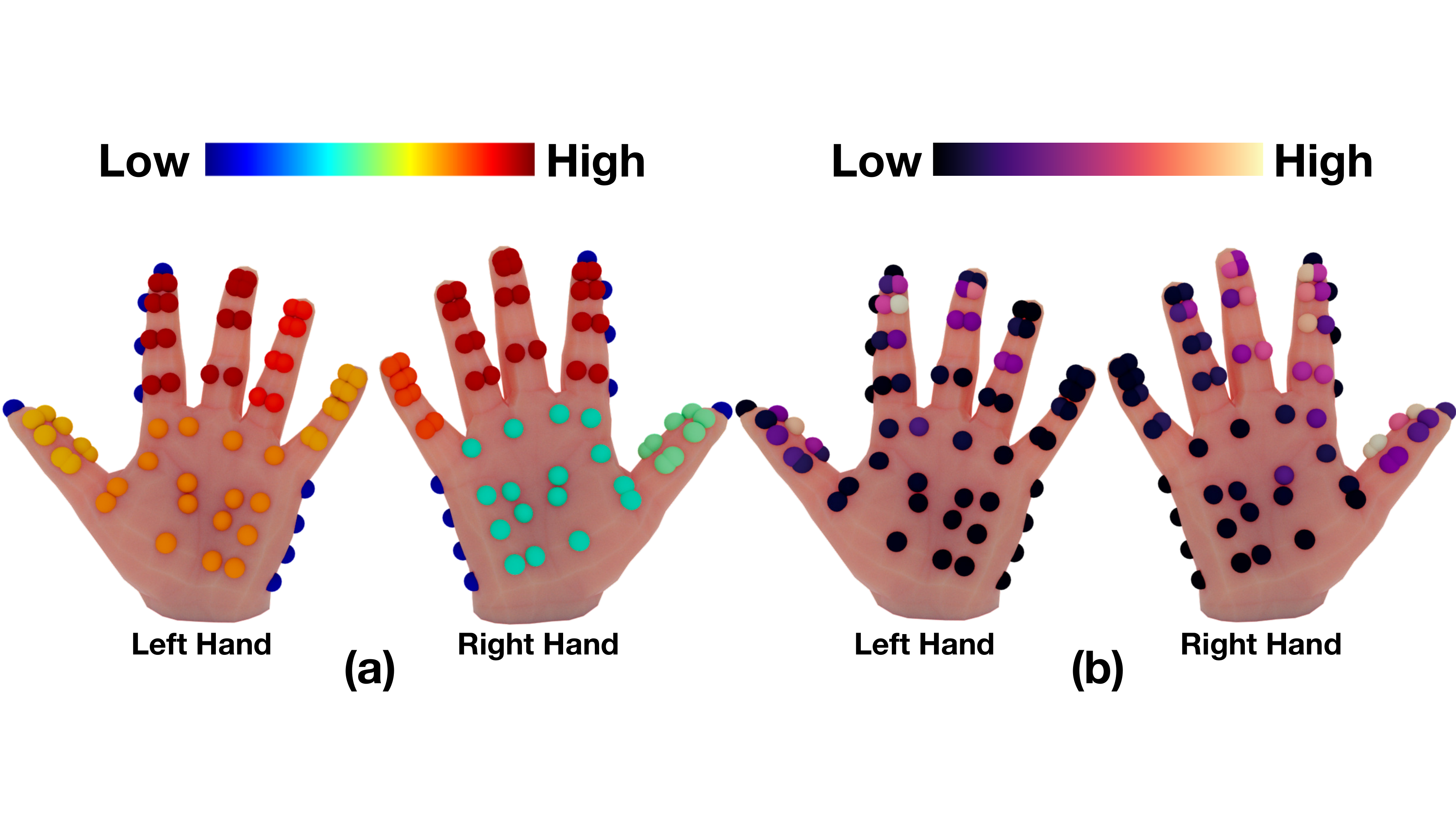}
\caption{Visualizations of the most commonly (a) contacted taxels and (b) applied pressures across all captures.}
\label{fig:tactilecoverage}
\end{figure}

\begin{figure}
\centering
\includegraphics[trim={0 0 0 0},clip,width=1.0\linewidth]{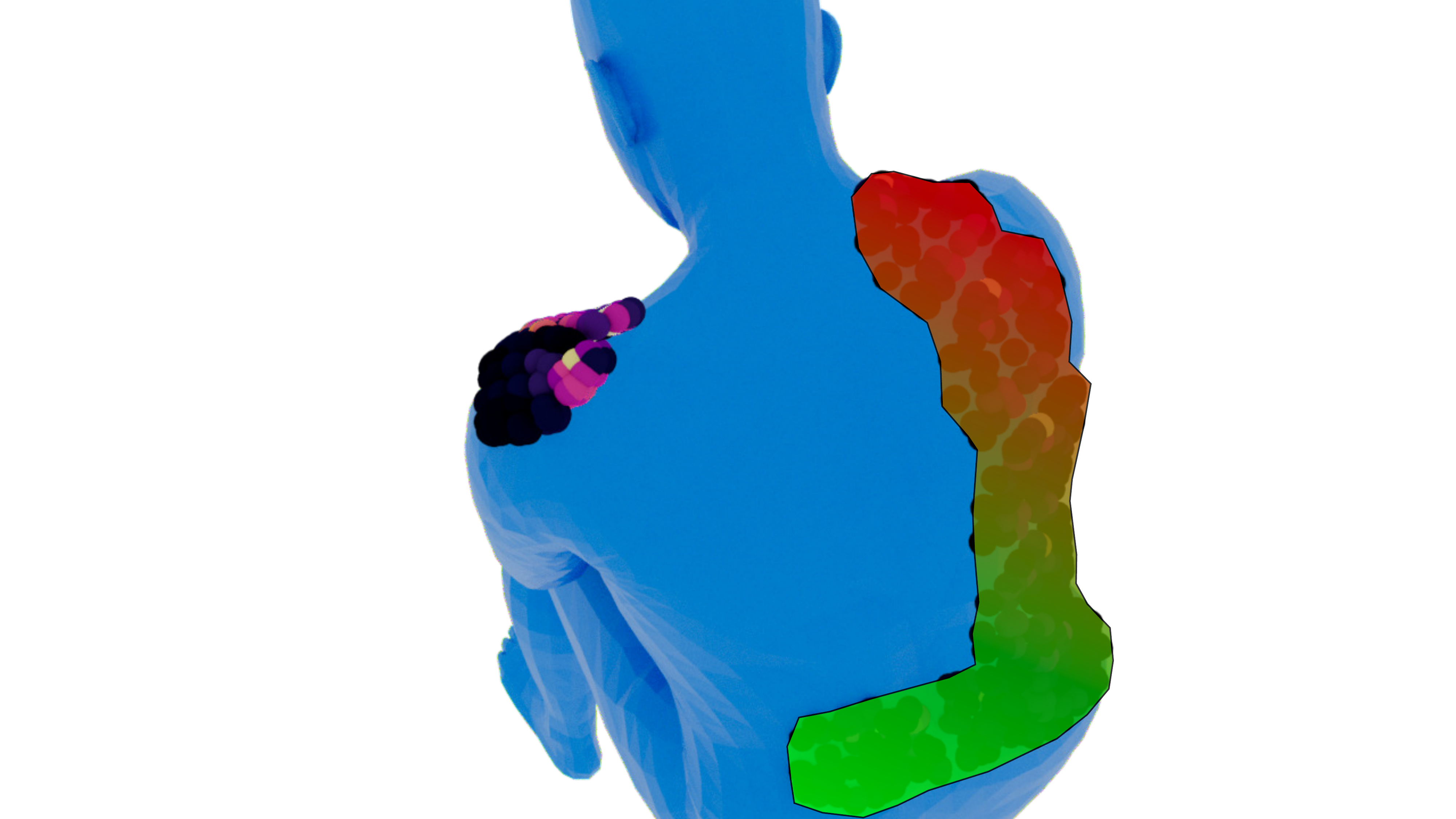}
\caption{Visualization of a bathing ``stroke" from the (green) bottom to the (red) top of the back. The subject is stabilized by a contact distribution at the left shoulder during the stroke.}
\label{fig:contactstroke}
\end{figure}

Figure~\ref{fig:bodycoverage} visualizes body coverage across all captures for two independent subjects, which we measure as all triangles that contain a contact point at any time during a bathing trajectory. Figure~\ref{fig:tactilecoverage} visualizes the most commonly contacted taxels on the left and right hands across all captures, as well as the distribution of forces across taxels. Figure~\ref{fig:contactstroke} illustrates a sample bathing ``stroke", or surface path integral of a contact distribution over time, of the right hand up the back.

\subsection{Quantitative Reconstruction Evaluation}

\begin{table}[h!]
\centering
\caption{$L2$ Contact distance metric comparison between MoSh++\cite{mahmood2019moshpp} and our reconstruction.}
\begin{tabular}{ |>{\centering\arraybackslash}M{1.5cm}|M{1.5cm}|M{1.5cm}|M{1.5cm}|}
    \hline
    \multicolumn{4}{|c|}{$L2$ Distance Comparisons (cm)} \\
    \hline
     & Median & Mean & St. Dev.\\
    \hline
    MoSh++ & 1.714 & 2.067 & 1.464\\
    \hline
    Ours & \textbf{0.536} & \textbf{0.837} & \textbf{0.842}\\
    \hline
\end{tabular}
\label{table:reconstructionmetrics}
\end{table}

\begin{figure}
\centering
\includegraphics[trim={0 0 0 0},clip,width=1.0\linewidth]{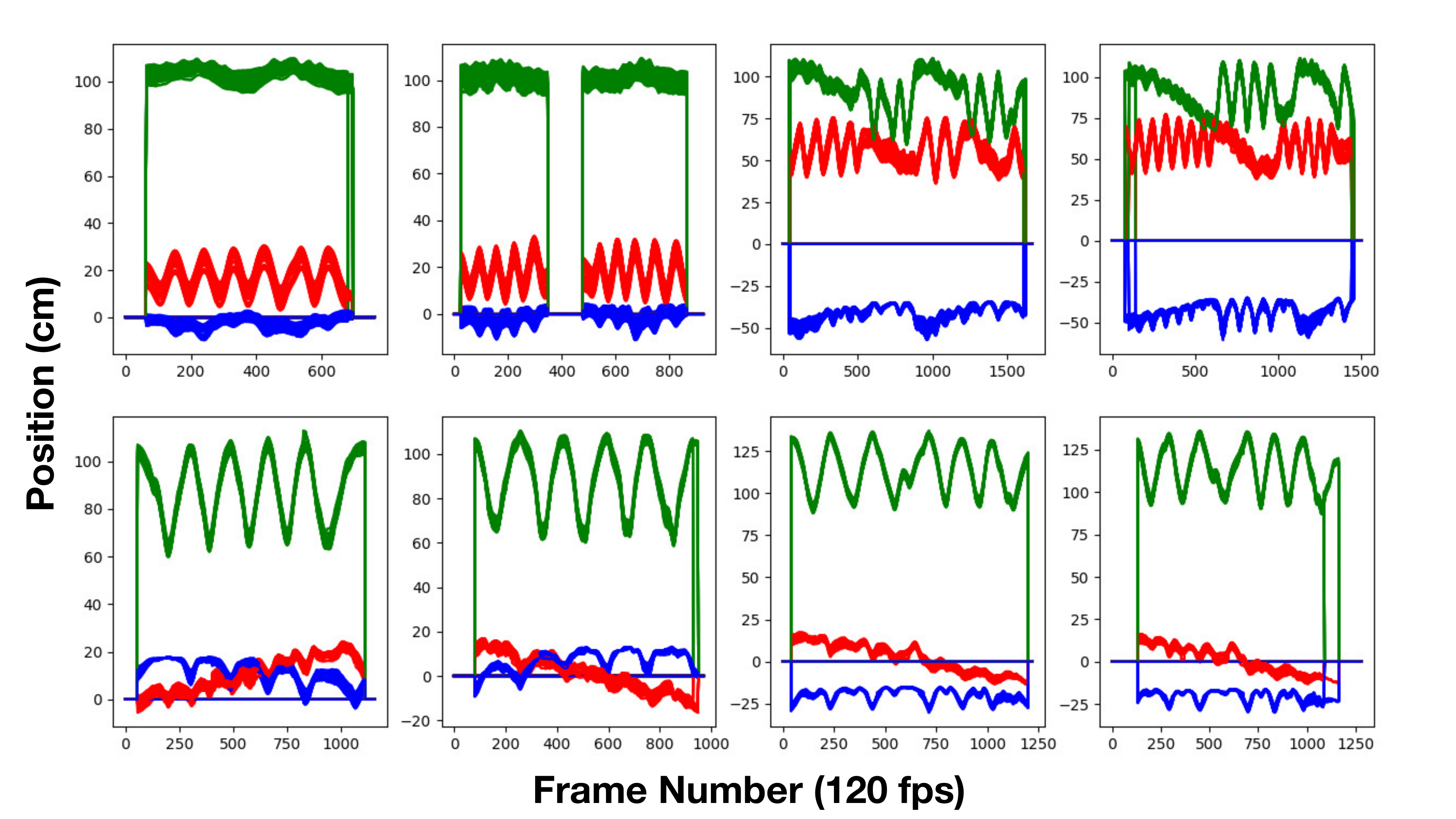}
\caption{Illustrations of 8 representative contact trajectories, separated into (red) X, (green) Y, and (blue) Z channels. All positions are in world coordinates. Each band is comprised of 65 individual contact point trajectories. Inactive taxels are zeroed out.}
\label{fig:contactcontinuity}
\end{figure}

We quantitatively evaluate the quality of our reconstruction pipeline using two metrics: $L2$ distances between estimated body contacts and glove taxels, and smoothness of body contact trajectories. Table~\ref{table:reconstructionmetrics} tabulates distance comparisons of our method against the original MoSh++~\cite{mahmood2019moshpp} reconstruction across 10 sample bathing demonstrations, while Figure~\ref{fig:contactcontinuity} plots the trajectories of all active wiping hand body-projected taxels across multiple back bathing demonstrations. We observe high spatiotemporal correlation in the projected taxel trajectories and that oscillations coincide with multiple back-and-forth wiping passes over back segments.

\subsection{Soft Hand Design and Control}

We perform and evaluate DexKit hand design using a subset of the dataset focusing on back bathing. We collect poses and contacts from the right hand in all subject captures to build $P$, and perform a sweep over the number of candidates in the diversity subset $S$ to determine the optimal sample size $|S|$. $|S|$ candidates are evaluated via MAE differentials between the original poses and their respective projections in $S$ obtained by solving Eq.~\ref{eq:qp}. We select $|S| = 50$ based on observing diminishing returns in the loss curve for larger values of $|S|$, and then choose $|S'| = 20$ based on a budget of six motors. We only consider poses in contact as candidates for $S$ and $S'$ and store all active taxels for each selected pose. $\vect{\beta}^*$ weights for all trajectory poses are computed by solving Eq.~\ref{eq:qp} using the resulting $\vect{M_{\vect{S'}}}$.

\begin{figure}
\centering
\includegraphics[trim={0 0 0 0},clip,width=1.0\linewidth]{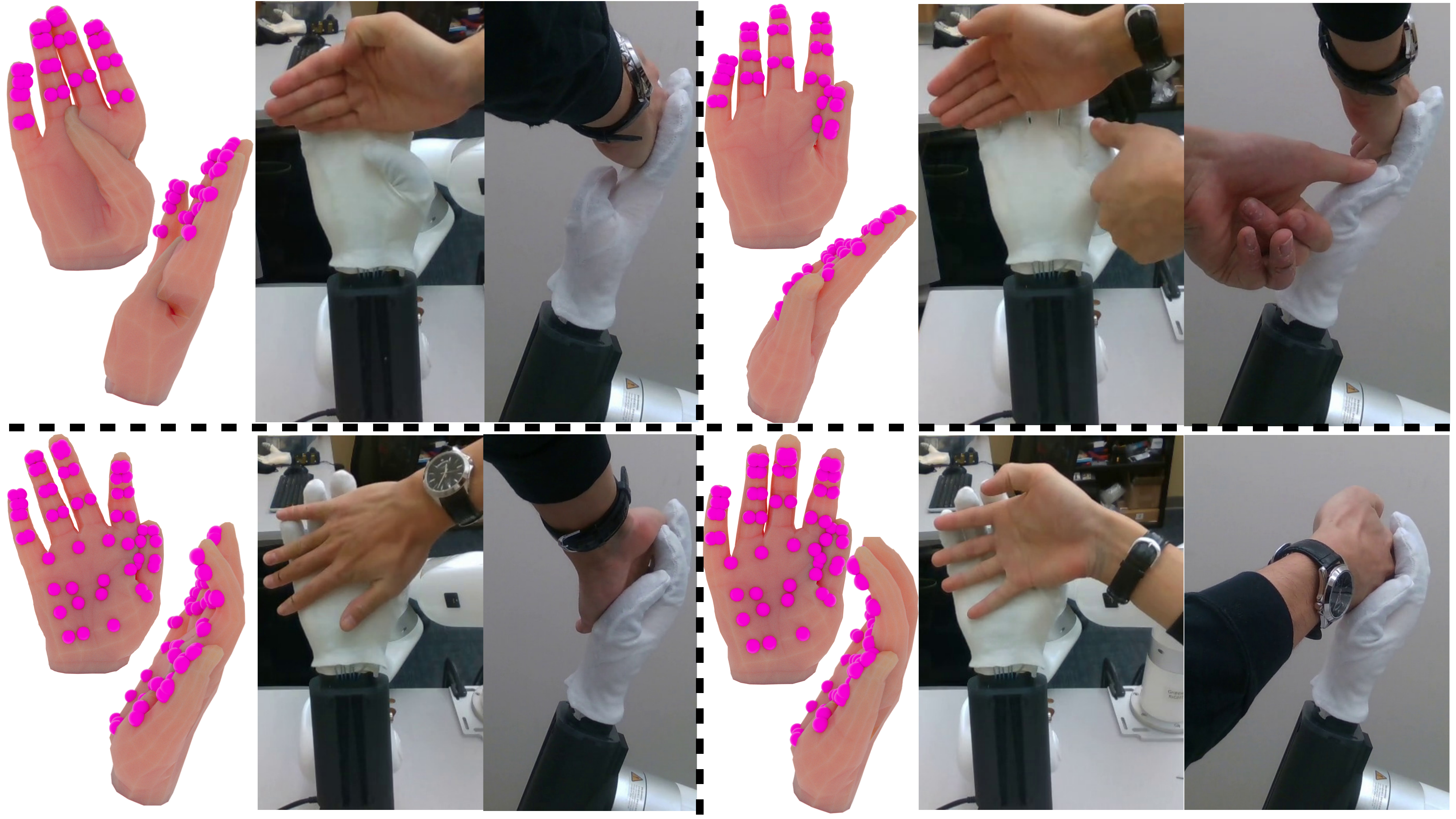}
\caption{Sample poses used to build $\vect{M_{\vect{\phi(S')}}}$. Active contacts in each pose (pink) are simulated by pressing against the hand.}
\label{fig:ethancalibration}
\end{figure}

We next build $\vect{M_{\vect{\phi(S')}}}$ and compute $\vect{\phi}$. We manually determine motor positions for each candidate pose without contact, and then adjust positions while pressing the hand at the regions of contact designated by the active taxels. Figure~\ref{fig:ethancalibration} illustrates several poses used during calibration.

\subsection{Contact Retargeting and Real World Deployment}

\begin{figure}
\centering
\includegraphics[trim={0 0 0 0},clip,width=1.0\linewidth]{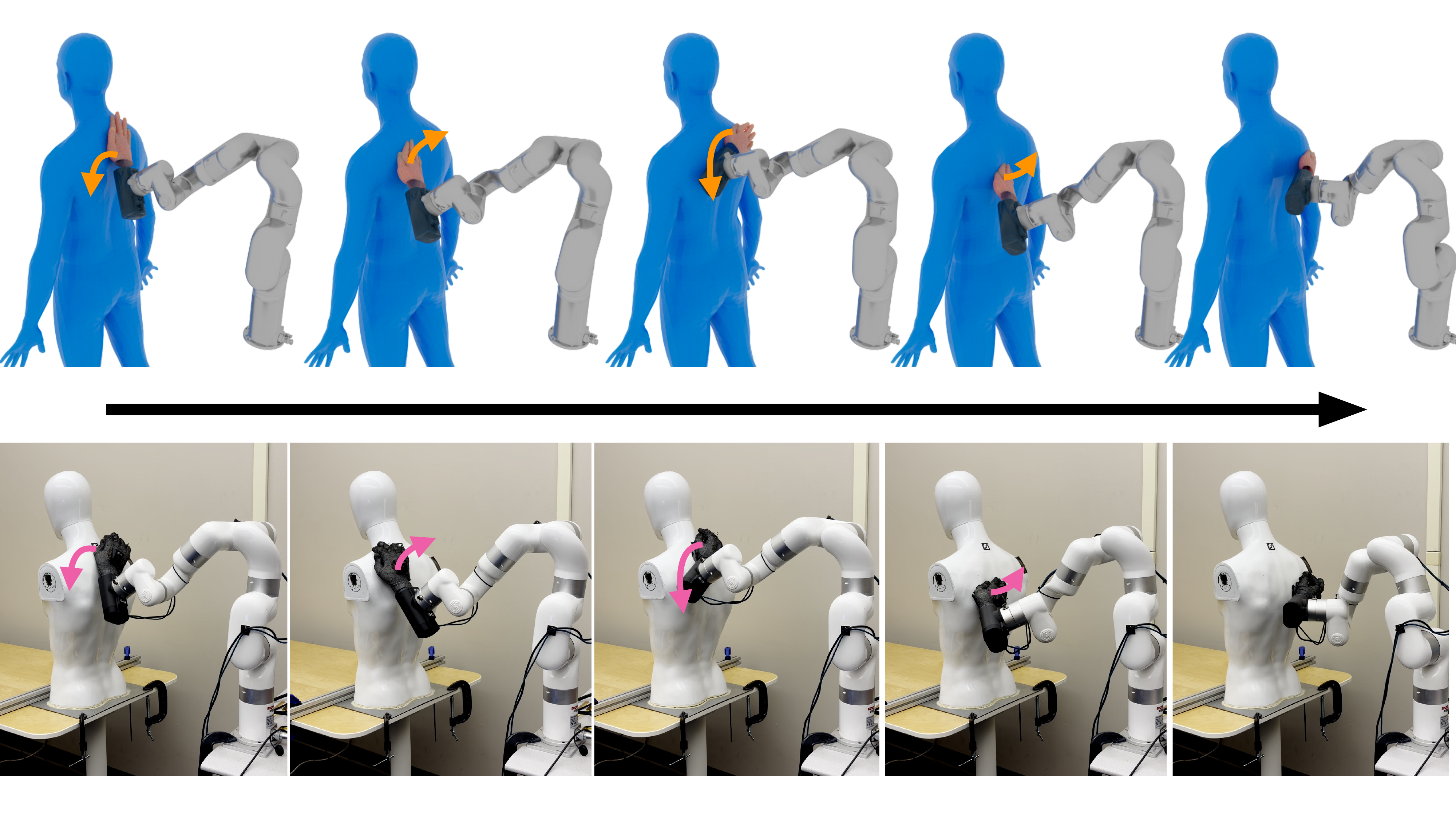}
\caption{Film strip of back bathing demonstration retargeted to the xArm 7 (top) in simulation using the clinician's shape-fitted hand as the end effector. The computed joint positions and pre-computed hand motor commands are (bottom) deployed on the real world setup. Orange and pink arrows illustrate subsequent movement directions.}
\label{fig:rolloutcomparison}
\end{figure}

\begin{figure}
\centering
\includegraphics[trim={0 0 0 0},clip,width=1.0\linewidth]{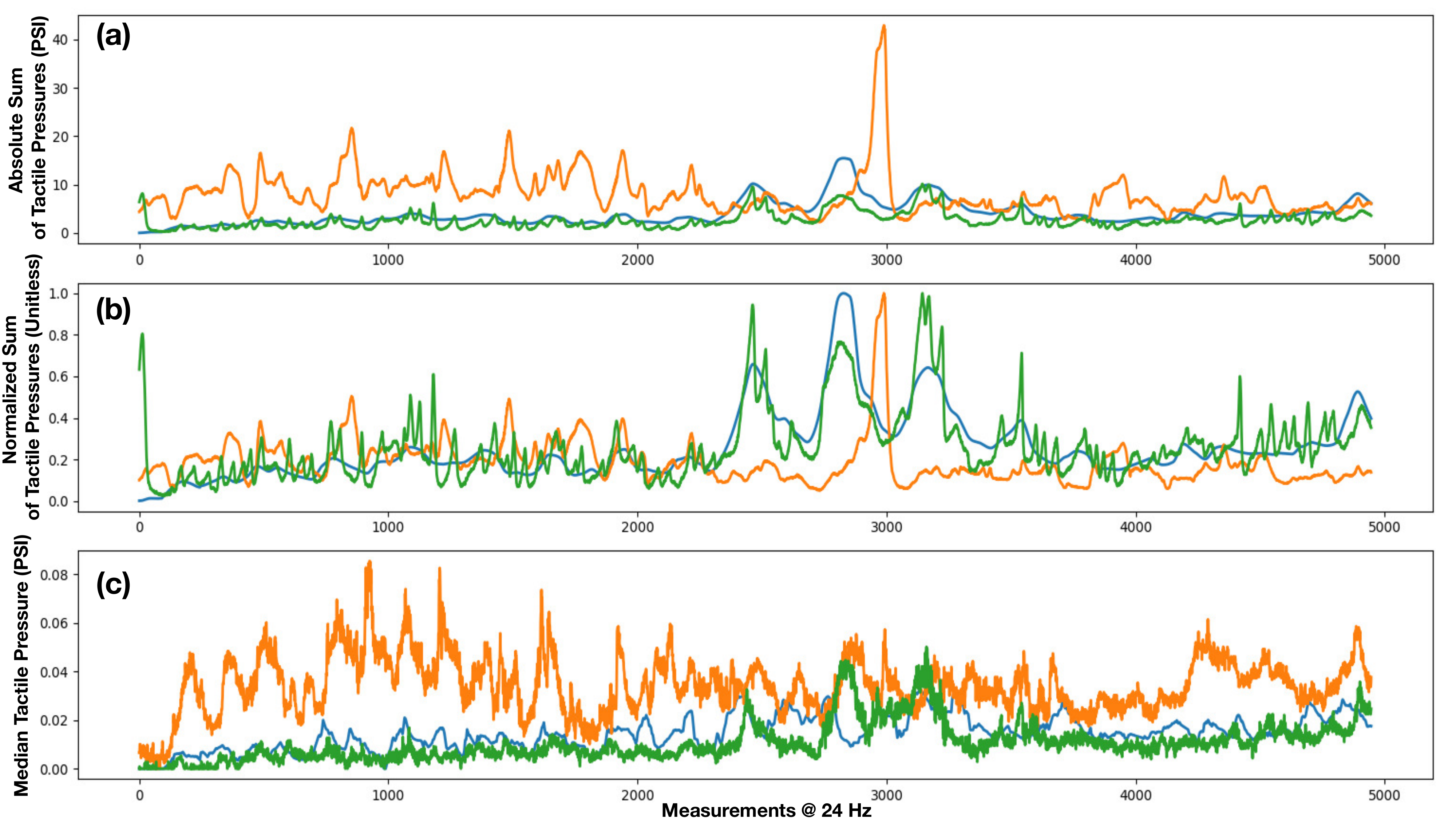}
\caption{Overlaid plots of the (blue) original human, (orange) robot open loop, and (green) robot closed loop tactile glove pressures during a back bathing task. We provide (a) absolute sums, (b) normalized sums, and (c) median intensities calculated across the full taxel array per time step. Note that the spike in pressure at the start of closed loop data collection is due to initialization noise rather than a consequence of trajectory execution.}
\label{fig:tactileplots}
\end{figure}

Figure~\ref{fig:rolloutcomparison} illustrates the xArm positions computed following a retargeted contact trajectory rollout, as well as deployment of that trajectory on our real world platform. 
$\vect{\psi}$ and $\vect{\phi}$ commands are deployed together in the control loop, and the original human demonstration is slowed from 120 to 6 Hz (20X). Complete results are available in the supplementary video. We provide additional Sim2Real alignment details in the appendix.

Figure~\ref{fig:tactileplots} compares the recorded glove pressure signals across open-loop rollout, closed-loop rollout, and the original human demonstration. As expected, the integrated tactile pressure for the open-loop rollouts is significantly higher than both the human demonstration and closed-loop results for the majority of the trajectory. The unsafe operating conditions indicated by these elevated pressure levels are further evidenced by the mannequin vibrations visible in the supplementary video. Closed-loop control substantially curtails such issues, as quantitatively evidenced in the plots and qualitatively by minimal mannequin vibrations in the video. Closed-loop pressures also reasonably track the original human demonstration, particularly when viewed on a normalized scale.

However, the closed-loop controller deviates from the human demonstration in absolute pressure and especially spatial distribution, as indicated by the comparison of median values. The distributional misalignment is an inherent trade-off of the controller design, as optimizing for an aggregate pressure sum does not necessitate parity in the underlying taxel-wise distribution. Furthermore, the attenuated dynamic range is a byproduct of necessary safety constraints in real world deployment. Specifically, we found that targeting a strict absolute pressure induces significant arm oscillations and mitigated this behavior by implementing tolerance bounds around the target instead. We also identified a failure mode where friction-induced finger curling during wiping motions could lead to mechanical damage. If curling occurs while the controller commands an inward corrective force, the system may drive the arm further into the mannequin in a futile attempt to reach a pressure target that the compromised finger geometry cannot achieve. To mitigate this, we implemented a temporal safety timeout for unreached pressure setpoints. Interestingly, this failure mode was absent in open-loop trials; because the controller advances through the trajectory regardless of feedback, the hand's passive compliance allowed fingers to recover from compression during lateral movements.

%% file: sections/discussion.tex
\section{Discussion}

Several key insights emerged during the reconstruction and transfer process. We outline our primary findings through the dataset, lessons learned, and current limitations below.

\subsection{Data Insights}

As strongly evidenced by video results and Figures~\ref{fig:contactstroke} and~\ref{fig:contactcontinuity}, bathing contact trajectories are \textit{highly} correlated, smooth, and dynamic. Modeling contacts as individual points, instantaneous static entities, or simple constraints, as is still commonly done by many long-standing analysis techniques~\cite{mason2001mechanics,ferrari1992planning,lynch2017modern} and modern simulators~\cite{todorov2012mujoco,makoviychuk2021isaac}, is highly limiting for representing our captured data. Contact models based on surface area~\cite{lakshmipathy2023contactedit,lakshmipathy2022contact} or volume~\cite{elandt2019pressure}—when paired with well-defined velocity parameters—offer significantly more promise for capturing the sustained interactions necessary for bathing.

The recorded tactile signals for mild pressure demonstrations, regardless of clinician or body part, are quite low (in multiple cases below the minimum sensitivity of 0.04 N) and thus largely indistinguishable from noise. We therefore strongly advise prioritizing high contact sensitivity in hardware design for low-pressure data acquisition, or alternatively focusing efforts on strong pressure captures.

Unsurprisingly, we found that signals between adjacent taxels tend to be highly correlated regardless of the task. We argue there is little value gained from high resolution sensing of regions unaffected by pose changes (e.g. the proximal phalanx of an individual finger). In contrast, hand shapes induced by pose changes, as well as hand shape in relation to local body shape, have a much more significant impact on pressure distributions. Therefore, we argue that thinking of sensor coverage in terms of \textit{regions between joints} is more valuable for bathing than intra-region resolution.

Interestingly, in captures where both hands were utilized, the majority of high pressure measurements came from the supporting hand rather than the wiping hand. This observation indicates that body stabilization requires substantially more force than wiping, and therefore suggests that assymetry may be a desirable characteristic of bimanual bathing systems. For example, the stabilizing arm and hand can utilize high torque, low gear-reduced actuators and sport limited DOFs, while the wiping hand can trade off power for speed.

Finally, we observed a \textit{substantial} amount of patient movement, both voluntary and in response to bathing forces. While our initial proof-of-concept transfers utilized a static mannequin, any system intended for human application \textit{must} account for constant body re-positioning.

\subsection{Lessons Learned}

While tactile information does help determine precise times and magnitude of contact, its utility is limited without context from motion and shape. Tactile data alone cannot determine \textit{between which bodies} contact is happening or \textit{why} forces change over time. Both issues made identifying true contact events between the hand and body difficult to automate. Multiple contact events between the thumb and thenar region of the palm, for example, resulted from gripping the wipe rather than applying body pressure. Drops in pressure could be attributed to either breaking contact or simply reducing contact force. Low, but increasing pressure signals could be actual contact events or drift noise. We found that while motion and shape data are vital for clarifying such events, they are not a total solution; a holistic analysis of all three channels together is still required to accurately interpret results.

We also found that compliance was \textit{of paramount importance} in arm-mounted bathing systems. Sustained, dense-contact interactions permit little room for trajectory mismatch errors, which are especially hard to avoid during sim2real rollouts. Although the Dexkit hand’s material compliance mitigated these issues during our experiments, we contend that incorporating compliance into arm control is essential for safe and effective interaction with human subjects.

Finally, we ultimately found that a flexor-only finger tendon configuration was insufficient to capture all necessary hand motions during rollouts. Passive compliance did not always allow the hand to escape friction-induced finger curling during wiping motions. Including dedicated extensor tendons would help mitigate such issues, but would be difficult to do under limited motor budgets.

\subsection{Limitations and Future Work}

Our system currently contains major current limitations. Our closed-loop strategy, for example, is highly limited by its use of tactile pressure sums rather than the distribution over individual taxels. The primary barrier to distributed force control is the significant discrepancy between human and Dexkit hand responses when subjected to similar pressure stimuli. While existing methods mitigate this issue through careful placement of sensors, selection of sensing modality, and calibration to align readings between similar human and robot trajectories~\cite{yin2025osmo,wi2026tactalign}, such approaches tend to be brittle and require significant engineering overhead. There remains an opportunity to more directly model the inherent discrepancies in hand materials. We propose that \textit{retargeting force distributions} to different geometries and material properties is a promising direction of future research.

Our approach's simulation-to-reality gap is substantial. While our force data is grounded purely in the real world, estimates of the xArm configurations and hand controller blend weights require substantial calibration due to mismatches in the mannequin vs. SMPL-X fitted model and MANO-reconstructed hand vs. Dexkit hand control spaces respectively. We also do not account for compliance in the Dexkit hand, reconstructed human body, or reconstructed human hand, which together can complicate accurately reproducing clinician-induced sensations beyond taxel observations.

Our data does not differentiate between normal and shear forces. This limitation can be attributed to our glove's capacitive sensing modality; alternatives such as visual~\cite{yuan2017gelsight} and magnetic~\cite{yin2025osmo} sensing can provide such measurements, albeit at the cost of form factor, lower SNR, reliability, or interpretability. While force direction will not impact our reconstruction pipeline, the information can be useful in online control scenarios such as helping more carefully manage applied frictional forces during sliding or escaping the identified curling failure mode.

Because of these and other limitations, our system is not yet ready for deployment on human subjects. Even with mechanical compliance, the primary bottleneck is perception. Specifically, no existing method achieves fully online human pose and shape estimation while maintaining the sub-frame latency required for live interaction, and our current assumption of a static body will not extend to a human subject. We argue that this component is vital both for retargeting contact trajectories from existing demonstrations and for enabling the system to react to human pose changes. Addressing this challenge could pave the way for the first whole-body bathing system capable of learning directly from human demonstrations.

%% file: sections/conclusion.tex
\section{Conclusion}

We have presented a straightforward, but effective framework for high fidelity capture and reconstruction of human bathing demonstrations, as well as methods to facilitate transfer of demonstrations across multiple levels of the robotics stack. Leveraging our framework, we have created the inaugural dataset of whole-body bathing demonstrations performed by trained clinicians on human subjects. Importantly, we have shown that utilizing contact regions as a key processing primitive effectively addresses many pain points in the reconstruction and transfer pipelines. We hope that the contributions of our methods, dataset, and code will spur further research in pHRI and ultimately make progress towards bringing useful and reliable robots into high-value clinical settings.

%% file: sections/acknowledgements.tex
\section*{ACKNOWLEDGMENTS}

We thank Justin C. Macey for assistance with data capture and cleanup, as well as Giorgio Becherini, Shashank Tripathi, and Alp\'{a}r Cseke for help in setting up the MoSh++ solver. Research reported in this publication was partially supported by the National Institute of Biomedical Imaging and Bioengineering of the National Institutes of Health under award number R01EB036842 and an award from the RAI Institute.

%% file: sections/appendix.tex
\appendix

We utilize this supplemental document to detail our real world setups, constraints, and design choices concerning the DexKit and xArm platforms. 

\subsection{DexKit Pose Selection Metrics}

\begin{figure}
\centering
\includegraphics[trim={0 0 0 0},clip,width=1.0\linewidth]{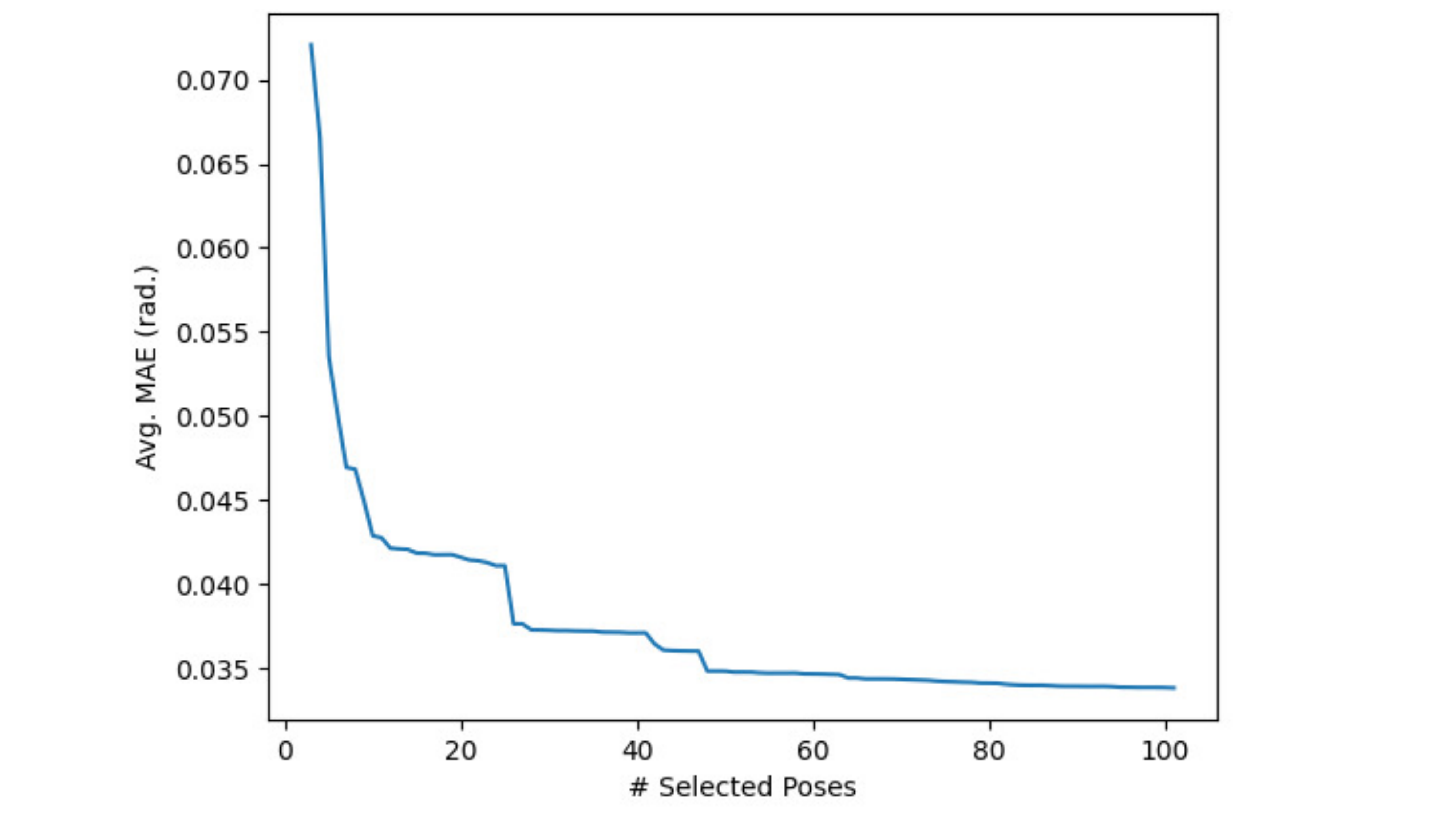}
\caption{Complete MAE loss curve over the parameter sweep of $|S|$.}
\label{fig:dexkitposesselection}
\end{figure}

Figure~\ref{fig:dexkitposesselection} illustrates the complete loss curve over a sweep of $|S| = 1$ to $|S|=100$. The size of all right hand back bathing poses $|P| = 9491$. While there are further dips in the curve at values $|S|>100$, and it is possible to achieve zero error by choosing $|S| = |P|$, we found that many poses were qualitatively repetitive. Errors of re-projection MAE losses computed using the finally constructed $M_{S'}$ were normally distributed with parameters $(\mu, \sigma) = (0.04477, 0.01928)$. Although these statistics suggest effective computed blend weights $\vect{\beta^*}$, we note that metrics concerning MANO hand poses are not necessarily strong measures for evaluating DexKit pose coverage. 

\subsection{Dexkit Motor Configuration and Tendon Routing}

\begin{figure}
\centering
\includegraphics[trim={0 5cm 0 5cm},clip,width=1.0\linewidth]{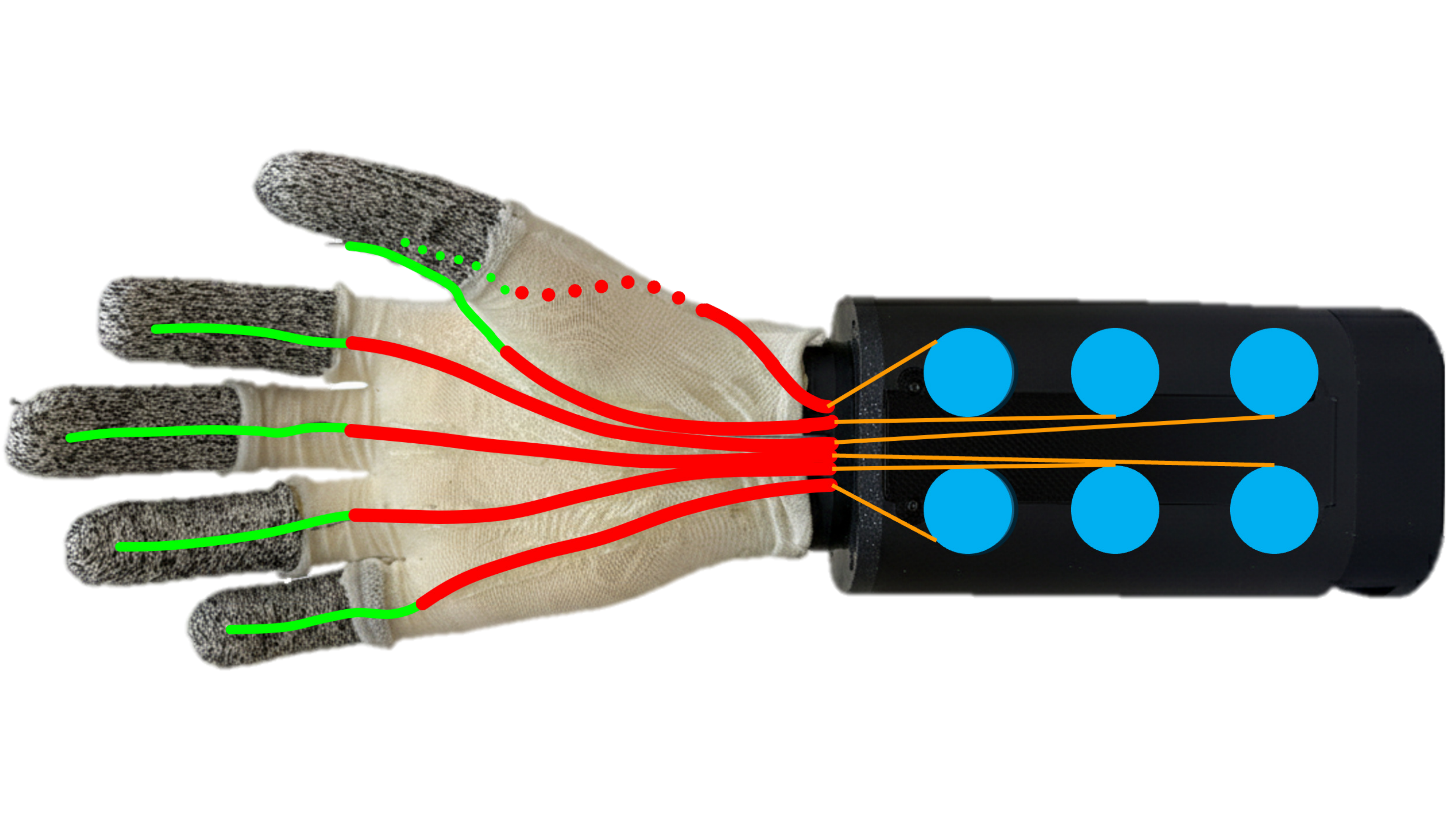}
\caption{The DexKit hand is comprised of a cast foam hand with a textile "skin". Tendons are sewn into the skin (green), and routed along the palm through captive PTFE sheaths (red). The hand shown has six tendons, five flexors on the fingers and thumb, as well as an adductor on the thumb which is routed along the back. The hand interfaces with six servo motors packaged in a housing block at the wrist (black). Tendons in the block (orange) route to pulleys (blue) on the motors which reel in the tendons to actuate the hand.}
\label{fig:dexkitschematics}
\end{figure}

Figure~\ref{fig:dexkitschematics} provides schematics of the final Dexkit motor and tendon configuration. We select a flexor-only configuration for the index, middle, ring, and pinky fingers, and an adduction routing for the thumb to pull in closer to the palm thenar region. We found the adduction routing more useful than a flexure routing because the flexed thumb interfered with the palm making contact with the mannequin during wiping the relatively flat back regions. The total system mass is roughly 0.25 kg, which is well within the xArm payload limit.

\subsection{Sim2Real Alignment Details}

\begin{figure}
\centering
\includegraphics[trim={0 0 0 0},clip,width=0.9\linewidth]{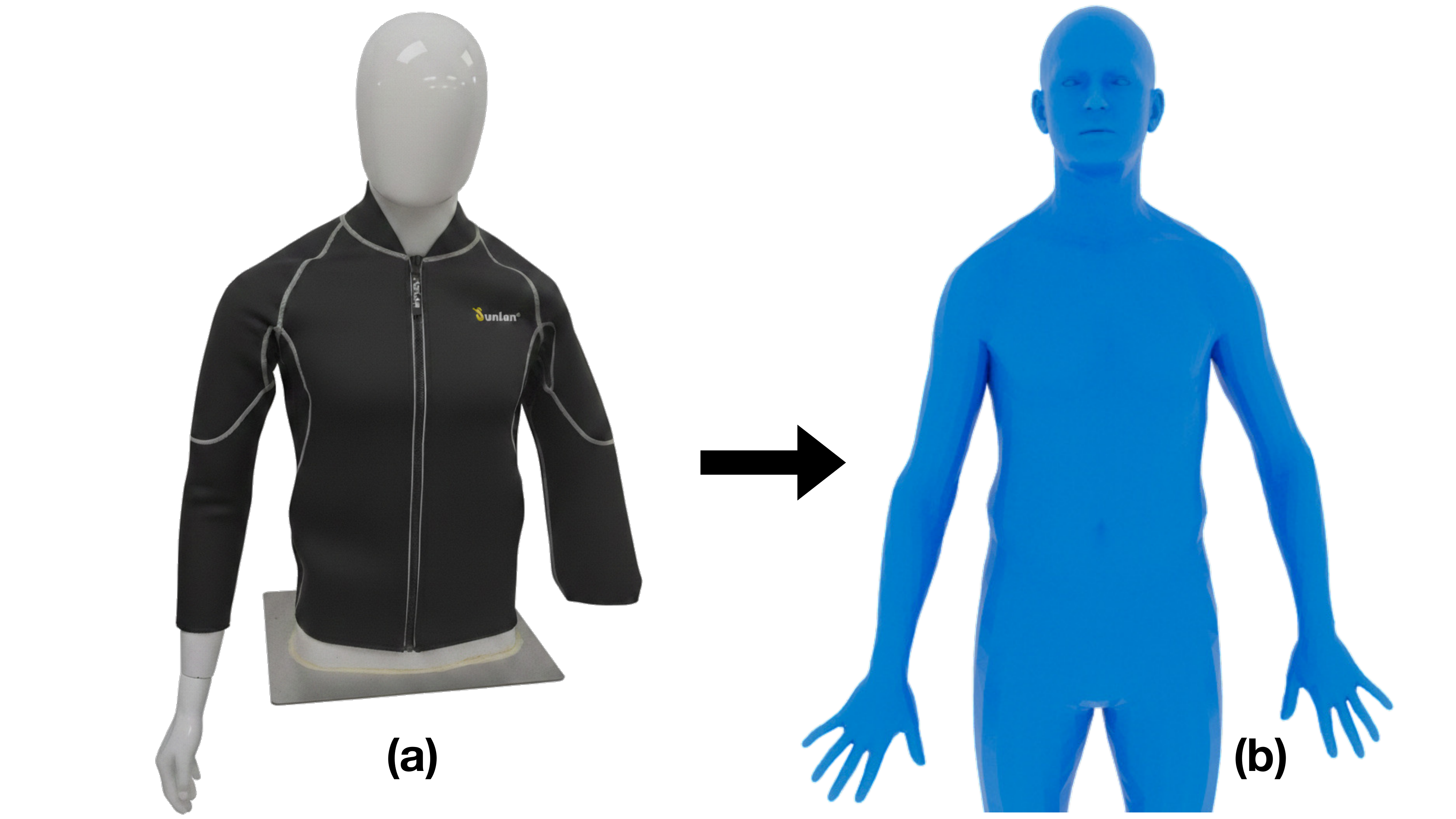}
\caption{Illustrations of the (a) mannequin input image and (b) extracted SMPL-X proxy mesh from TRAM~\cite{wang2024tram}.}
\label{fig:tramreconstruction}
\end{figure}

\begin{figure}
\centering
\includegraphics[trim={0 0 0 0},clip,width=1.0\linewidth]{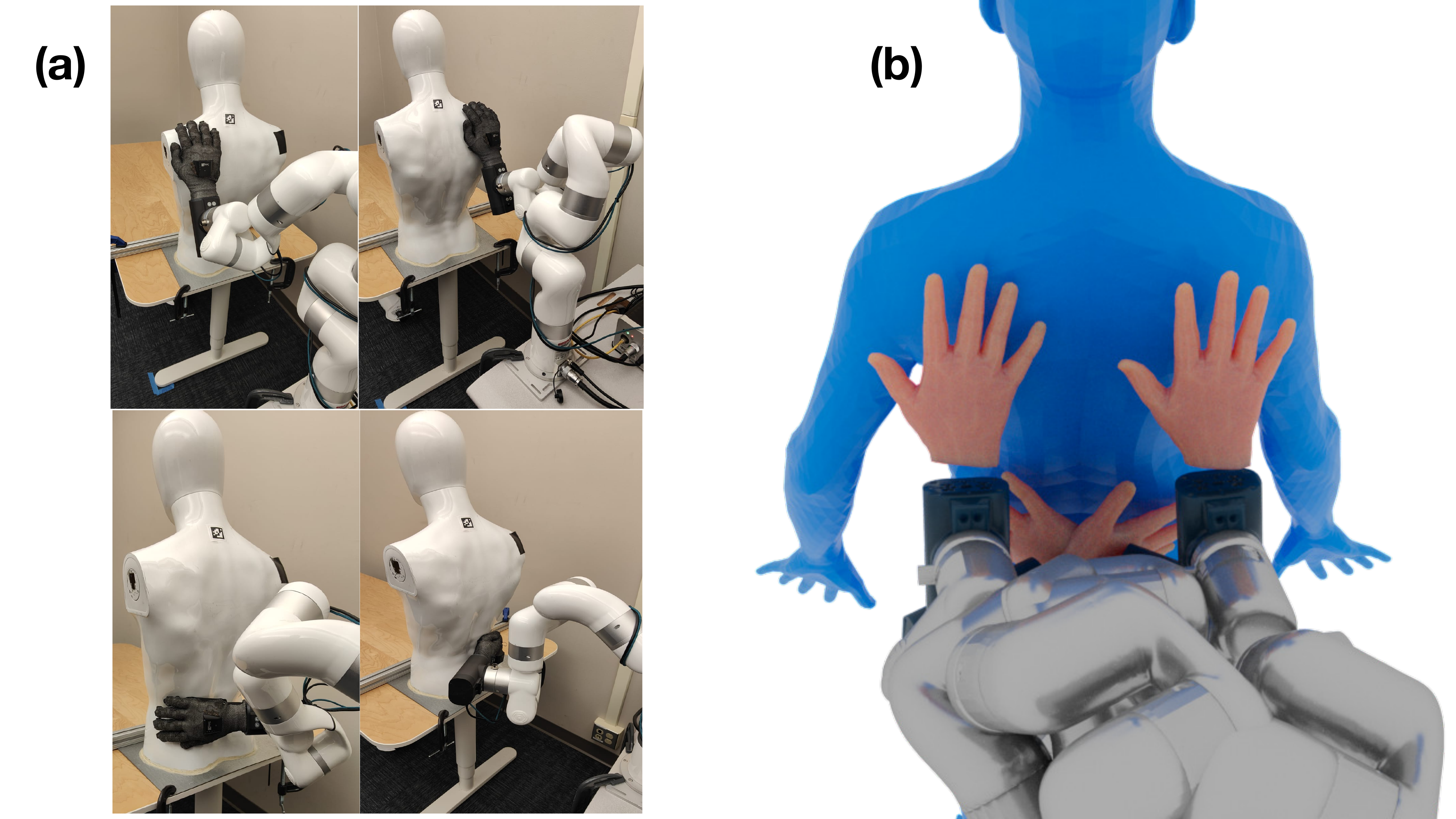}
\caption{(a) xArm calibration poses collected in the real world are (b) superimposed for mannequin positioning in simulation.}
\label{fig:calibrationimage}
\end{figure}

Figure~\ref{fig:tramreconstruction} depicts the mannequin image and TRAM-reconstructed SMPL-X body shape, averaged across 10 frames of video. Notably, we found that attaching at least one arm and placing clothing on the mannequin was necessary because the model otherwise failed to identify the structure as a human. We only utilize the baseline body shape parameters, ignoring blend shape corrections and root pose estimates.

Instead, we estimate the root position through real-world calibration. We manually adjust the xArm into four distinct poses to estimate the ``boundaries" of the mannequins back, which include the left and right shoulders as well as the left and right hip segments. We then manually align the mannequin proxy in simulation relative to the corresponding xArm positions, and note that we intentionally offset the proxy mesh slightly further from the hands. This accounts for shape discrepancies and ensures the arm biases toward overreaching rather than falling short. Figure~\ref{fig:calibrationimage} illustrates the process.

%% file: references.bib
@article{adeniji2025feel,
    title={Feel the Force: Contact-Driven Learning from Humans},
    author={Adeniji, A. and Chen, Z. and Liu, V. and Pattabiraman, V. and Bhirangi, R. and Haldar, S. and Abbeel, P. and Pinto, L.},
    journal={arXiv preprint arXiv:2506.01944},
    year={2025}
}

@article{baxter1997capacitive,
  title={Capacitive sensors},
  author={Baxter, L. K.},
  journal={IEEE Design and Applications},
  year={1997}
}

@inproceedings{brahmbhatt2019contactgrasp,
    title={Contactgrasp: Functional multi-finger grasp synthesis from contact},
    author={Brahmbhatt, S. and Handa, A. and Hays, J. and Fox, D.},
    booktitle={2019 IEEE/RSJ International Conference on Intelligent Robots and Systems},
    pages={2386--2393},
    year={2019},
    organization={IEEE}
}

@inproceedings{cheng2022contact,
    title={Contact mode guided motion planning for quasidynamic dexterous manipulation in 3d},
    author={Cheng, X. and Huang, E. and Hou, Y. and Mason, M. T.},
    booktitle={International Conference on Robotics and Automation},
    pages={2730--2736},
    year={2022},
    organization={IEEE}
}

@inproceedings{christen2022d,
    title={D-grasp: Physically plausible dynamic grasp synthesis for hand-object interactions},
    author={Christen, S. and Kocabas, M. and Aksan, E. and Hwangbo, J. and Song, J. and Hilliges, O.},
    booktitle={IEEE/CVF Conference on Computer Vision and Pattern Recognition},
    pages={20577--20586},
    year={2022}
}

@article{czuba2012ergonomic,
    title={Ergonomic and safety risk factors in home health care: Exploration and assessment of alternative interventions},
    author={Czuba, L. R. and Sommerich, C. M. and Lavender, S. A.},
    journal={Work},
    volume={42},
    number={3},
    pages={341--353},
    year={2012}
}

@article{darragh2015musculoskeletal,
    title={Musculoskeletal discomfort, physical demand, and caregiving activities in informal caregivers},
    author={Darragh, A. R. and Sommerich, C. M. and Lavender, S. A. and Tanner, K. J. and Vogel, K. and Campo, M.},
    journal={Journal of Applied Gerontology},
    volume={34},
    number={6},
    pages={734--760},
    year={2015}
}

@article{delpreto2022actionsense,
    title={Actionsense: A multimodal dataset and recording framework for human activities using wearable sensors in a kitchen environment},
    author={DelPreto, J. and Liu, C. and Luo, Y. and Foshey, M. and Li, Y. and Torralba, A. and Matusik, W. and Rus, D.},
    journal={Advances in Neural Information Processing Systems},
    volume={35},
    pages={13800--13813},
    year={2022}
}

@inproceedings{dometios2017real,
    title={Real-time end-effector motion behavior planning approach using on-line point-cloud data towards a user adaptive assistive bath robot},
    author={Dometios, A. C. and Papageorgiou, X. S. and Arvanitakis, A. and Tzafestas, C. S. and Maragos, P.},
    booktitle={IEEE/RSJ International Conference on Intelligent Robots and Systems},
    pages={5031--5036},
    year={2017}
}

@article{dunlop1997disability,
    title={Disability in activities of daily living: patterns of change and a hierarchy of disability.},
    author={Dunlop, D. D. and Hughes, S. L. and Manheim, L. M.},
    journal={American journal of public health},
    volume={87},
    number={3},
    pages={378--383},
    year={1997},
    publisher={American Public Health Association}
}

@inproceedings{elandt2019pressure,
    title={A pressure field model for fast, robust approximation of net contact force and moment between nominally rigid objects},
    author={Elandt, R. and Drumwright, E. and Sherman, M. and Ruina, A.},
    booktitle={IEEE/RSJ International Conference on Intelligent Robots and Systems},
    pages={8238--8245},
    year={2019},
    organization={IEEE}
}

@inproceedings{erickson2019multidimensional,
    title={Multidimensional capacitive sensing for robot-assisted dressing and bathing},
    author={Erickson, Z. and Clever, H. M. and Gangaram, V. and Turk, G. and Liu, C. K. and Kemp, C. C.},
    booktitle={IEEE International Conference on Rehabilitation Robotics},
    pages={224--231},
    year={2019}
}

@inproceedings{fan2023arctic,
    title={ARCTIC: A dataset for dexterous bimanual hand-object manipulation},
    author={Fan, Z. and Taheri, O. and Tzionas, D. and Kocabas, M. and Kaufmann, M. and Black, M. J. and Hilliges, O.},
    booktitle={IEEE/CVF Conference on Computer Vision and Pattern Recognition},
    pages={12943--12954},
    year={2023}
}

@article{fang2025dexop,
    title={Dexop: A device for robotic transfer of dexterous human manipulation},
    author={Fang, H. S. and Romero, B. and Xie, Y. and Hu, A. and Huang, B. R. and Alvarez, J. and Kim, M. and Margolis, G. and Anbarasu, K. and Tomizuka, M. and others},
    journal={arXiv preprint arXiv:2509.04441},
    year={2025}
}

@inproceedings{ferrari1992planning,
    title={Planning optimal grasps},
    author={Ferrari, C. and Canny, J. and others},
    booktitle={Proceedings., 1992 IEEE International Conference on Robotics and Automation, 1992.},
    volume={3},
    pages={2290--2295},
    year={1992},
    organization={IEEE}
}

@article{ghosh1996computational,
    title={Computational aspects of the maximum diversity problem},
    author={Ghosh, J. B.},
    journal={Operations Research Letters},
    volume={19},
    number={4},
    pages={175--181},
    year={1996}
}

@article{gill2006epidemiology,
    title={The epidemiology of bathing disability in older persons},
    author={Gill, T. M and Guo, A. and Allore, H. G.},
    journal={Journal of the American Geriatrics Society},
    volume={54},
    number={10},
    pages={1524--1530},
    year={2006}
}

@inproceedings{huang2022soft,
    title={Soft tactile contour following for robot-assisted wiping and bathing},
    author={Huang, I. and Chow, D. and Bajcsy, R.},
    booktitle={IEEE/RSJ International Conference on Intelligent Robots and Systems},
    pages={7797--7802},
    year={2022}
}

@article{katz1963studies,
    title={Studies of illness in the aged: the index of ADL: a standardized measure of biological and psychosocial function},
    author={Katz, S. and Ford, A. B. and Moskowitz, R. W. and Jackson, B. A. and Jaffe, M. W.},
    journal={Journal of the American Medical Association},
    volume={185},
    number={12},
    pages={914--919},
    year={1963}
}

@inproceedings{king2010towards,
  title={Towards an assistive robot that autonomously performs bed baths for patient hygiene},
  author={King, C. H. and Chen, T. L. and Jain, A. and Kemp, C. C.},
  booktitle={IEEE/RSJ International Conference on Intelligent Robots and Systems},
  pages={319--324},
  year={2010}
}

@inproceedings{king2025dexkit,
    author    = {King, J. P. and  Ahluwalia, H. and Heredia-Marin, I. B. and McClary, K. and Zuo, E. and Baratz, M. and Colgate, J. E. and Clark, J. and {Orta Martinez}, M. and Pollard, N. S.},
    booktitle = {Proceedings of the 19th International Symposium on Experimental Robotics},
    title     = {DexKit: A Hardware Platform for Multi-Finger Dexterous Robotics},
    year      = {2025}
}

@article{kuo1993analyzing,
    title={Analyzing and modeling the maximum diversity problem by zero-one programming},
    author={Kuo, C. C. and Glover, R. and Dhir, K. S.},
    journal={Decision Sciences},
    volume={24},
    number={6},
    pages={1171--1185},
    year={1993}
}

@inproceedings{lakshmipathy2022contact,
  title={Contact transfer: A direct, user-driven method for human to robot transfer of grasps and manipulations},
  author={Lakshmipathy, A. and Bauer, D. and Bauer, C. and Pollard, N. S.},
  booktitle={2022 International Conference on Robotics and Automation (ICRA)},
  pages={6195--6201},
  year={2022},
  organization={IEEE}
}

@article{lakshmipathy2023contactedit,
    title={Contact Edit: Artist Tools for Intuitive Modeling of Hand-Object Interactions},
    author={Lakshmipathy, A. S. and Feng, N. and Lee, Y. X. and Mahler, M. and Pollard, N. S.},
    journal={ACM Transactions on Graphics},
    year={2023},
    volume={42},
    number={4},
    articleno={45}
}

@article{lakshmipathy2025kinematic,
    title={Kinematic motion retargeting for contact-rich anthropomorphic manipulations},
    author={Lakshmipathy, A. S. and Hodgins, J. K. and Pollard, N. S.},
    journal={ACM Transactions on Graphics},
    volume={44},
    number={2},
    pages={1--20},
    year={2025}
}

@inproceedings{liu2022characterization,
    title={Characterization of a Meso-Scale Wearable Robot for Bathing Assistance},
    author={Liu, F. and Patil, V. and Erickson, Z. and Temel, Z.},
    booktitle={IEEE International Conference on Robotics and Biomimetics},
    pages={2146--2152},
    year={2022},
    organization={IEEE}
}

@article{luo2024adaptive,
    title={Adaptive tactile interaction transfer via digitally embroidered smart gloves},
    author={Luo, Y. and Liu, C. and Lee, Y. J. and DelPreto, J. and Wu, K. and Foshey, M. and Rus, D. and Palacios, T. and Li, Y. and Torralba, a. and others},
    journal={Nature Communications},
    volume={15},
    number={1},
    pages={868},
    year={2024}
}

@book{lynch2017modern,
  title={Modern robotics},
  author={Lynch, K. M. and Park, F. C.},
  year={2017},
  publisher={Cambridge University Press}
}

@inproceedings{madan2024rabbit,
    title={Rabbit: A robot-assisted bed bathing system with multimodal perception and integrated compliance},
    author={Madan, R. and Valdez, S. and Kim, D. and Fang, S. and Zhong, L. and Virtue, D. T. and Bhattacharjee, T.},
    booktitle={Proceedings of the 2024 ACM/IEEE international conference on human-robot interaction},
    pages={472--481},
    year={2024}
}

@book{magee2013orthopedic,
    title={Orthopedic Physical Assessment},
    author={Magee, D. J.},
    year={2013},
    publisher={Elsevier Health Sciences}
}

@inproceedings{magnenat1989joint,
    title={Joint-dependent local deformations for hand animation and object grasping},
    author={Magnenat-Thalmann, N. and Laperri{\`e}re, R. and Thalmann, D.},
    booktitle={Graphics Interface},
    pages={26--33},
    year={1989}
}

@inproceedings{mahmood2019moshpp,
    title={AMASS: Archive of Motion Capture as Surface Shapes},
    author={Mahmood, N. and Ghorbani, N. and Troje, N. F. and Pons-Moll, G. and Black, M. J.},
    booktitle = {IEEE International Conference on Computer Vision},
    pages={5442--5451},
    year={2019}
}

@article{makoviychuk2021isaac,
  title={Isaac gym: High performance gpu-based physics simulation for robot learning},
  author={Makoviychuk, V. and Wawrzyniak, L. and Guo, Y. and Lu, M. and Storey, K. and Macklin, M. and Hoeller, D. and Rudin, N. and Allshire, A. and Handa, A. and others},
  journal={arXiv preprint arXiv:2108.10470},
  year={2021}
}

@article{mandi2025dexmachina,
  title={Dexmachina: Functional retargeting for bimanual dexterous manipulation},
  author={Mandi, Z. and Hou, Y. and Fox, D. and Narang, Y. and Mandlekar, A. and Song, S.},
  journal={arXiv preprint arXiv:2505.24853},
  year={2025}
}

@article{mao2025visuo,
    title={Visuo-Acoustic Hand Pose and Contact Estimation},
    author={Mao, Y. and Yoo, U. and Yao, Y. and Syed, S. N. and Bondi, L. and Francis, J. and Oh, J. and Ichnowski, J.},
    journal={arXiv preprint arXiv:2508.00852},
    year={2025}
}

@book{mason2001mechanics,
    title={Mechanics of robotic manipulation},
    author={Mason, M. T.},
    year={2001},
    publisher={MIT press}
}

@article{pan2025spider,
    title={SPIDER: Scalable Physics-Informed Dexterous Retargeting},
    author={Pan, C. and Wang, C. and Qi, H. and Liu, Z. and Bharadhwaj, H. and Sharma, A. and Wu, T. and Shi, G. and Malik, J. and Hogan, F.},
    journal={arXiv preprint arXiv:2511.09484},
    year={2025}
}

@article{pang2023global,
    title={Global planning for contact-rich manipulation via local smoothing of quasi-dynamic contact models},
    author={Pang, T. and Suh, H. J. T. and Yang, L. and Tedrake, R.},
    journal={IEEE Transactions on robotics},
    volume={39},
    number={6},
    pages={4691--4711},
    year={2023},
    publisher={IEEE}
}

@inproceedings{pavlakos2019expressive,
    title={Expressive body capture: 3d hands, face, and body from a single image},
    author={Pavlakos, G. and Choutas, V. and Ghorbani, N. and Bolkart, T. and Osman, A. A. A. and Tzionas, D. and Black, M. J},
    booktitle={IEEE/CVF Conference on Computer Vision and Pattern Recognition},
    pages={10975--10985},
    year={2019}
}

@misc{ppsTactileGlove,
    author = {{Pressure Profile Systems}},
    title = {{TactileGlove}: {Hand} Pressure and Force Measurement},
    year = {2026},
    howpublished = {\url{https://pressureprofile.com/body-pressure-mapping/tactile-glove}},
    note = {Accessed: 2026-01-26}
}

@misc{3MGripTape,
    author = {{3M™}},
    title = {{Gripping Material TB641}},
    year = {2026},
    howpublished = {\url{https://www.3m.com/3M/en_US/p/d/b40069682/}},
    note = {Accessed: 2026-01-26}
}

@article{romero2017mano,
    title = {Embodied Hands: Modeling and Capturing Hands and Bodies Together},
    author = {Romero, J. and Tzionas, D. and Black, M. J.},
    journal = {ACM Transactions on Graphics},
    volume = {36},
    number = {6},
    year = {2017}
}

@misc{sawhneyFCPW,
    author = {Sawhney, R.},
    title = {FCPW: Fastest Closest Points in the West},
    version = {1.0},
    year = {2021},
    url = {https://github.com/rohan-sawhney/fcpw}
}

@article{sivakumar2022robotic,
  title={Robotic telekinesis: Learning a robotic hand imitator by watching humans on youtube},
  author={Sivakumar, A. and Shaw, K. and Pathak, D.},
  journal={arXiv preprint arXiv:2202.10448},
  year={2022}
}

@article{song2025opentouch,
    title={OPENTOUCH: Bringing Full-Hand Touch to Real-World Interaction},
    author={Song, Y. R. and Li, J. and Fu, R. and Murphy, D. and Zhou, K. and Shiv, R. and Li, Y. and Xiong, H. and Owens, C. E. and Du, Y. and others},
    journal={arXiv preprint arXiv:2512.16842},
    year={2025}
}

@article{sundaram2019learning,
    title={Learning the signatures of the human grasp using a scalable tactile glove},
    author={Sundaram, S. and Kellnhofer, P. and Li, Y. and Zhu, J. Y. and Torralba, A. and Matusik, W.},
    journal={Nature},
    pages={698--702},
    year={2019}
}

@inproceedings{taheri2020grab,
    title={GRAB: A dataset of whole-body human grasping of objects},
    author={Taheri, O. and Ghorbani, N. and Black, M. J. and Tzionas, D.},
    booktitle={European conference on computer vision},
    pages={581--600},
    year={2020},
    organization={Springer}
}

@inproceedings{todorov2012mujoco,
    title={Mujoco: A physics engine for model-based control},
    author={Todorov, E. and Erez, T. and Tassa, Y.},
    booktitle={IEEE/RSJ International conference on intelligent robots and systems},
    pages={5026--5033},
    year={2012}
}

@inproceedings{turpin2022grasp,
    title={Grasp’d: Differentiable contact-rich grasp synthesis for multi-fingered hands},
    author={Turpin, D. and Wang, L. and Heiden, E. and Chen, Y. C. and Macklin, M. and Tsogkas, S. and Dickinson, S. and Garg, A.},
    booktitle={European Conference on Computer Vision},
    pages={201--221},
    year={2022},
    organization={Springer}
}

@misc{ufactoryXarm7,
    author = {{UFACTORY}},
    title = {{UFACTORY xArm 7 Robotic Arm}},
    year = {2026},
    howpublished = {\url{https://www.ufactory.us/product/ufactory-xarm-7}},
    note = {Accessed: 2026-01-26}
}

@misc{viconV16Specs,
    author = {{Vicon Motion Systems}},
    title = {{V16} Camera Specifications - {Vantage} Documentation},
    year = {2026},
    howpublished = {\url{https://help.vicon.com/space/Vantage/15041618/V16+camera+specifications}},
    note = {Accessed: 2026-01-26}
}

@inproceedings{wang2024tram,
    title={TRAM: Global Trajectory and Motion of 3D Humans from in-the-wild Videos},
    author={Wang, Y. and Wang, Z. and Liu, L. and Daniilidis, K.},
    booktitle={European Conference on Computer Vision},
    pages={467--487},
    year={2024}
}

@article{wi2026tactalign,
    title={TactAlign: Human-to-Robot Policy Transfer via Tactile Alignment},
    author={Wi, Y. and Yin, J. and Xiang, E. and Sharma, A. and Malik, J. and Mukadam, M. and Fazeli, N. and Hellebrekers, T.},
    journal={arXiv:2602.13579},
    year={2026}
}

@inproceedings{wu2022saga,
    title={Saga: Stochastic whole-body grasping with contact},
    author={Wu, Y. and Wang, J. and Zhang, Y. and Zhang, S. and Hilliges, O. and Yu, F. and Tang, S.},
    booktitle={European Conference on Computer Vision},
    pages={257--274},
    year={2022},
    organization={Springer}
}

@article{xing2025taccap,
    title={TacCap: A Wearable FBG-Based Tactile Sensor for Seamless Human-to-Robot Skill Transfer},
    author={Xing, C. and Li, H. and Wei, Y. L. and Ren, T. A. and Tu, T. and Lin, Y. and Schumann, E. and Zheng, W. S. and Cutkosky, M. R.},
    journal={arXiv preprint arXiv:2503.01789},
    year={2025}
}

@article{yin2025osmo,
    title={OSMO: Open-Source Tactile Glove for Human-to-Robot Skill Transfer},
    author={Yin, J. and Qi, H. and Wi, Y. and Kundu, S. and Lambeta, M. and Yang, W. and Wang, C. and Wu, T. and Malik, J. and Hellebrekers, T.},
    journal={arXiv preprint arXiv:2512.08920},
    year={2025}
}

@inproceedings{zlatintsi2018multimodal,
    title={Multimodal signal processing and learning aspects of human-robot interaction for an assistive bathing robot},
    author={Zlatintsi, A. and Rodomagoulakis, I. and Koutras, P. and Dometios, A. C. and Pitsikalis, V. and Tzafestas, C. S. and Maragos, P.},
    booktitle={IEEE International Conference on Acoustics, Speech and Signal Processing},
    pages={3171--3175},
    year={2018}
}

@article{yuan2017gelsight,
    title={Gelsight: High-resolution robot tactile sensors for estimating geometry and force},
    author={Yuan, W. and Dong, S. and Adelson, E. H.},
    journal={Sensors},
    volume={17},
    number={12},
    pages={2762},
    year={2017}
}

@inproceedings{wu2024gello,
  title={Gello: A general, low-cost, and intuitive teleoperation framework for robot manipulators},
  author={Wu, P. and Shentu, Y. and Yi, Z. and Lin, X. and Abbeel, P.},
  booktitle={2024 IEEE/RSJ International Conference on Intelligent Robots and Systems},
  pages={12156--12163},
  year={2024}
}
